\documentclass[10pt,twocolumn,letterpaper]{article}

 \usepackage{cvpr}              

\definecolor{cvprblue}{rgb}{0.21,0.49,0.74}
\definecolor{citypink}{rgb}{0.85,0.2,0.53}
\definecolor{citypurple}{HTML}{6926AA}
\definecolor{pipelineblue}{HTML}{1F4EA8}
\definecolor{pipelineorange}{HTML}{F7941D}
\definecolor{pipelinegreen}{HTML}{4C9445}

\usepackage[pagebackref,breaklinks,colorlinks,allcolors=citypurple]{hyperref}
\usepackage[table]{xcolor}
\usepackage[most]{tcolorbox}
\usepackage{booktabs}   
\usepackage{multirow}   
\usepackage[table]{xcolor} 
\newtcbox{\highlighttext}[3]{%
  arc=4pt, %
  colback=#2, %
  boxrule=0pt, %
  left=0.3pt, right=0pt, top=-1.5pt, bottom=-1.5pt,
  nobeforeafter,
  tcbox raise base,
  enhanced,
  fontupper=\textcolor{#1}{#3}%
}

\newcommand{\rkone}[1]{\cellcolor{citypurple!55}#1}
\newcommand{\rktwo}[1]{\cellcolor{citypurple!30}#1}
\newcommand{\rkthree}[1]{\cellcolor{citypurple!13}#1}

\def\confName{CVPR}
\def\confYear{2026}

\usepackage[utf8]{inputenc} 
\usepackage[T1]{fontenc}    
\usepackage{graphicx}
\usepackage{placeins}
\usepackage{wrapfig}
\usepackage{url}            
\usepackage{booktabs}       
\usepackage{multirow}
\usepackage{tabularx}
\usepackage{array}
\usepackage{amsmath}
\usepackage{amsfonts}       
\usepackage{nicefrac}       
\usepackage{microtype}      
\usepackage{cuted}
\usepackage{stfloats}       
\usepackage{multicol}       
\usepackage{xcolor}         

\usepackage[most]{tcolorbox} \usepackage{capt-of} 
\newtcolorbox{AIBox}[1]{ enhanced, width=\linewidth, colback=gray!4, colframe=black!65, colbacktitle=black!65, coltitle=white, title={#1}, fonttitle=\bfseries\small, boxrule=0.55pt, arc=1mm, left=1.2mm, right=1.2mm, top=1.0mm, bottom=1.0mm, before skip=4pt, after skip=2pt, before upper={\setlength{\parindent}{0pt}} }

\newenvironment{PromptText}{%
    \par
    \scriptsize
    \ttfamily
    \raggedright
    \sloppy
    \setlength{\parindent}{0pt}%
    \setlength{\parskip}{0pt}%
    \obeylines
}{%
    \par
}

\graphicspath{{ECCV/}}

\title{FreeStyle: Free Control for Style–Content Dual-Reference Generation from Community LoRA Mining}

\author{
Jinghong Lan$^{1,2*}$~~
Wei Cheng$^{2*}$~~
Yunuo Chen$^{2}$~~
Ziqi Ye$^{1}$~~
Peng Xing$^{2}$~~
Yixiao Fang$^{2}$~~
Rui Wang$^{2}$\\
Yufeng Yang$^{2}$~~
Xuanyang Zhang$^{2}$~~
Xianfang Zeng$^{2}$~~
Difan Zou$^{4}$~~
Gang Yu$^{2\ddag}$~~
Chi Zhang$^{3\ddag}$ \\ [0.5em]
$^{1}$ Fudan University~~
$^{2}$ StepFun~~
$^{3}$ Westlake University\quad
$^{4}$ University of Hong Kong \\[0.8em]
\textcolor{citypurple}{\normalsize
\raisebox{-0.2\height}{\includegraphics[height=0.5cm]{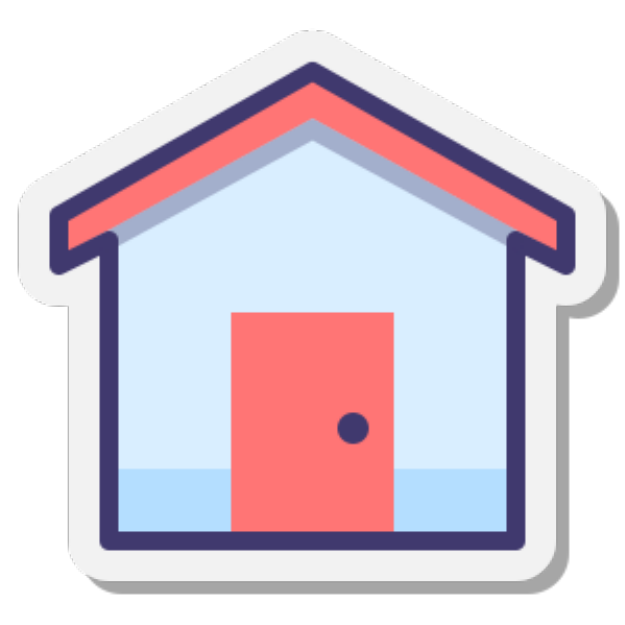}}~{\href{https://blue2giant.github.io/FreeStyle/}{\textbf{Project Page}}}
\quad
\raisebox{-0.2\height}{\includegraphics[height=0.5cm]{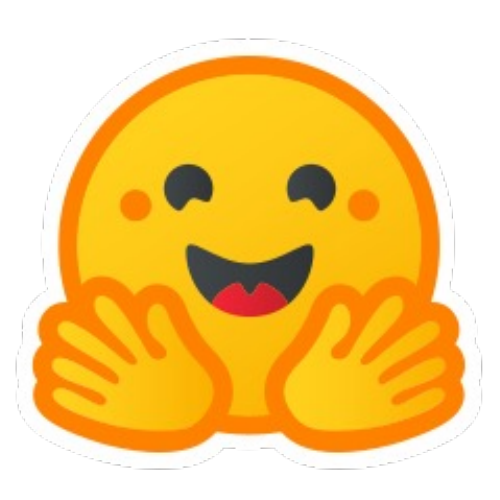}}~{\href{https://huggingface.co/datasets/Blue2Giant/FreeStyle_Dataset}{\textbf{Dataset}}}
\quad
\raisebox{-0.2\height}{\includegraphics[height=0.5cm]{files/huggingface_logo.pdf}}~{\href{https://huggingface.co/datasets/Blue2Giant/FreeStyle_Bench}{\textbf{Benchmark}}}
\quad
\raisebox{-0.2\height}{\includegraphics[height=0.5cm]{files/huggingface_logo.pdf}}~{\href{https://huggingface.co/Blue2Giant/FreeStyle_Checkpoint}{\textbf{Weights}}}
\quad
\raisebox{-0.2\height}{\includegraphics[height=0.5cm]{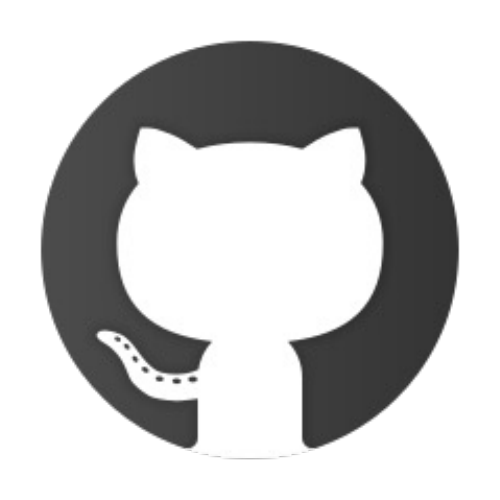}}~{\href{https://github.com/Blue2Giant/FreeStyle}{\textbf{Code}}}
}
}

\begin{document}

 \twocolumn[{
   \renewcommand\twocolumn[1][]{#1}
   \maketitle
   \begin{center}
     \vspace{-5mm}
     \includegraphics[width=\linewidth]{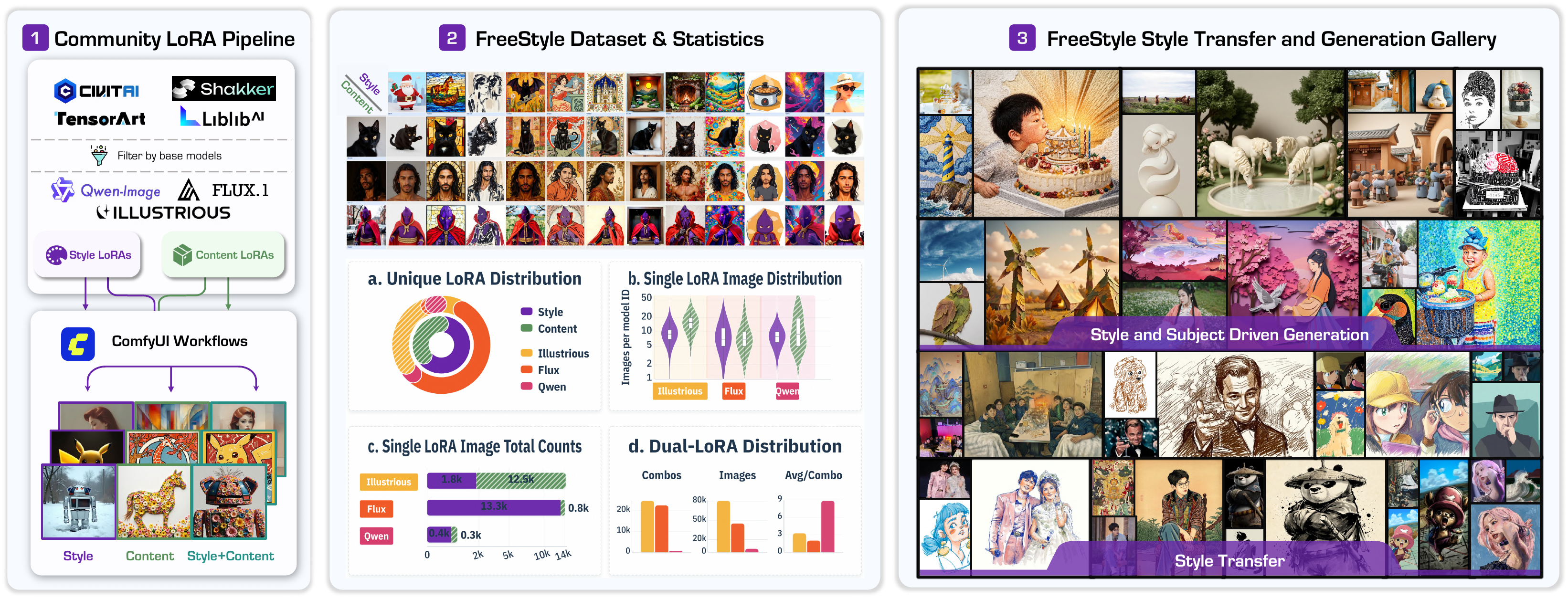}
     \vspace{-5mm}
    \captionof{figure}{\small \textbf{Overview of FreeStyle}. \highlighttext{white}{citypurple}{\textbf{1}}~~We collect community-created style and content LoRAs from multiple platforms and automatically compose them through standardized workflows.
\highlighttext{white}{citypurple}{\textbf{2}}~~The resulting \textit{FreeStyle} dataset contains diverse style--content image triplets spanning multiple base models, artistic styles, and subject categories.
\highlighttext{white}{citypurple}{\textbf{3}}~~FreeStyle enables both style transfer and style-subject controllable image generation across a broad range of visual domains.}
     \label{fig:triplet_show}
   \end{center}
 }]

\let\thefootnote\relax\footnotetext{* Equal contribution. \ddag~Corresponding authors. }

\begin{abstract}

Style- and content-dual-reference generation aims to synthesize an image that preserves the structure and semantics of a content reference while adopting the style of a separate style reference.
Despite recent progress, this setting remains challenging because models must balance content fidelity, style alignment, and instruction following while avoiding semantic leakage from the style reference.
A key bottleneck is the lack of large-scale triplet data with clean content-style separation and broad long-tail style coverage.
In this work, we propose FreeStyle, a scalable dual-reference generation framework based on community LoRA mining.
We treat community LoRAs as compositional anchors for style and content, and design a rigorous generation and filtering pipeline to construct large-scale content–style dual-reference triplets across multiple base models.
To address content leakage, we adopt a two-stage curriculum with stage-specific disentanglement mechanisms: an attention-level enrichment constraint that suppresses style-reference leakage in the style-transfer stage, and a frequency-aware RoPE modulation strategy that targets positional-correspondence-based leakage in the harder dual-reference stage.
We also introduce a benchmark covering both style-reference and dual-reference generation, with evaluations on style similarity, content preservation, aesthetics, instruction following, and VLM-based verification. The benchmark incorporates a style-invariant Content Alignment Score (CAS) and introduces a  VLM-based Verification Score for evaluating generation reliability and potential cross-reference leakage.
Extensive experiments show that our model achieves a strong balance among style alignment, content preservation, and leakage suppression.
\end{abstract}

\section{Introduction}

Reference-based image generation has become an effective paradigm for controllable visual synthesis, built upon the rapid progress of diffusion models~\cite{ho2020ddpm,song2021ddim,dhariwal2021diffusion,ho2022cfg} and large-scale text-to-image systems~\cite{rombach2022ldm,podell2024sdxl,ramesh2022dalle2,saharia2022imagen,nichol2022glide}.
Existing methods typically use external images to provide either style guidance or content reference, but style- and content-dual-reference generation remains a more challenging setting: given a content reference, a separate style reference, and a text instruction, the model must synthesize an image that preserves the structure of the content reference while adopting the visual style of the style reference.
This requires the model to jointly balance content fidelity, style alignment, and instruction following.

A key bottleneck is the lack of large-scale triplet data with clean content-style separation and broad style coverage.
Existing data construction pipelines either cover only a narrow set of styles, require costly manual curation, or produce triplets with imperfect content-style separation.
In this work, we propose FreeStyle, a scalable framework for dual-reference generation based on community LoRA mining.
Our key observation is that community LoRAs collectively provide a naturally curated and parameterized collection of visual concepts: each LoRA is typically trained around a coherent style, subject, or theme, and the whole community covers a broad spectrum of artistic styles and content categories.
We use these LoRAs as compositional anchors for both style and content, and design a rigorous mining, generation, and filtering pipeline to construct large-scale Style-Reference (SRef) and Content-Reference (CRef) triplets with broad long-tail style coverage and clean content-style separation.

Beyond data construction, another central difficulty is content-style disentanglement.
The style image often contains not only visual attributes such as color palette, texture, and brushwork, but also semantic content such as objects and layouts.
When used as a conditioning signal, these semantic elements can leak into the generated image, causing unwanted hallucinations or structural distortions.
We find that this leakage manifests through different mechanisms depending on the generation setting.
In style-reference generation, leakage primarily arises from disproportionate attention allocation to style-reference tokens during late denoising steps.
In the harder dual-reference setting, where a content reference absorbs much of the model's attention, leakage instead occurs through local positional correspondence encoded in high-frequency RoPE components, which enables patch-level copying from the style image.

To address these distinct failure modes, we adopt a two-stage training curriculum with stage-specific disentanglement mechanisms.
In Stage~1, the model is trained on style-transfer data to build robust style-reference generation, with an attention-level enrichment constraint that suppresses abnormal style-reference dominance while preserving style richness.
In Stage~2, we introduce dual-reference data and apply frequency-aware RoPE modulation to the style-reference branch, suppressing high-frequency positional components that encourage copying while amplifying low-frequency components that preserve global stylistic structure.
The two stages target complementary leakage pathways and together achieve a better balance among style alignment, content fidelity, and leakage control.

We also introduce a benchmark for systematically evaluating both style-reference and dual-reference generation.
Beyond standard feature-based metrics, we adopt the style-invariant Content Alignment Score (CAS) from CSGO~\cite{xing2024csgo} to measure structural agreement after factoring out style, and propose a  VLM-based Verification Score that separately quantifies style-transfer reliability and content preservation under potential cross-reference leakage.
This evaluation protocol exposes the trade-offs that a single aggregate score would obscure.

Our contributions are summarized as follows:
\begin{itemize}
    \item We propose FreeStyle, a scalable dual-reference generation framework that mines community LoRAs as compositional anchors for style and content, enabling large-scale construction of SRef and CRef triplets across multiple base models with broad long-tail style coverage.
    \item We introduce a systematic benchmark for both style-reference and dual-reference generation, together with a VLM-based Verification Score that separately evaluates style-transfer reliability and content preservation under potential cross-reference leakage.
    \item We propose a two-stage training strategy with stage-specific disentanglement: an attention-level enrichment constraint for style-reference generation, and frequency-aware RoPE modulation for dual-reference generation, each targeting a distinct leakage mechanism.
\end{itemize}

\section{Related Work}\label{sec:related}

\subsection{Reference-Based Generation and Stylization}

Reference-based generation controls image synthesis using external visual examples.
For content control, ControlNet~\cite{zhang2023controlnet} and T2I-Adapter~\cite{mou2024t2iadapter} inject spatial signals such as edges, depth, or poses into diffusion models; IP-Adapter~\cite{ye2023ipadapter} uses decoupled cross-attention for image-reference conditioning; and personalization methods adapt models to user-provided concepts via fine-tuning, low-rank adaptation, or token optimization~\cite{ruiz2023dreambooth,gal2023textual,kumari2023customdiffusion,hu2022lora,sohn2023styledrop}.

Style-reference generation further requires extracting visual attributes---color, texture, brushwork---from a reference image.
Neural style transfer has evolved from optimization-based and feed-forward formulations~\cite{gatys2016style,johnson2016perceptual,ulyanov2016texture,dumoulin2017learned} through arbitrary zero-shot stylization~\cite{huang2017adain,li2017wct,chenschmidt2016styleswap,sheng2018avatarnet} to attention- and transformer-based methods~\cite{park2019sanet,liu2021adaattn,deng2022stytr2}, reversible flows~\cite{an2021artflow}, learned linear transforms~\cite{li2019linear}, feature-distribution and optimal-transport matching~\cite{zhang2022efdm,kolkin2019strotss}, contrastive objectives~\cite{wu2022ccpl,chen2021ieconst}, and wavelet-based designs~\cite{yoo2019wct2,luan2017deepphoto}.
More recently, diffusion models enable training-free stylization via shared or swapped attention~\cite{hertz2024stylealigned,jeong2024visualstyle}, inversion-based methods~\cite{chung2024styleid,zhang2023inst}, and LoRA merging~\cite{shah2024ziplora,frenkel2024blora}, while GAN-based generators~\cite{zhu2017cyclegan,goodfellow2014gan,karras2019stylegan,karras2020stylegan2} offer domain-specific style control.
However, a style reference often carries semantic content that leaks into the output.
StyleAlign~\cite{hertz2024stylealigned}, InstantStyle~\cite{wang2024instantstyle}, DEADiff~\cite{qi2024deadiff}, and CleanStyle~\cite{feng2026cleanstyle} address this through disentanglement or purification strategies; CSGO~\cite{xing2024csgo} proposes end-to-end content-style composition; and EasyRef~\cite{zong2024easyref} supports generalized group-image references via multimodal LLMs.
These methods advance individual axes of reference-based generation, but most do not target large-scale supervision for explicit content-style dual-reference generation.

\subsection{Content and Style Dual-Reference Generation}

In the dual-reference setting, the model must simultaneously preserve the content reference, adopt the style reference, and follow a text instruction---requiring multi-image fusion and content-style disentanglement beyond single-reference or text-guided editing methods~\cite{meng2022sdedit,hertz2023prompt2prompt,brooks2023instructpix2pix}.
General-purpose editors such as GPT-Image~1.5~\cite{openai2025gptimage15}, Nano Banana Pro~\cite{google2026nanobananapro}, Qwen-Image-Edit~\cite{qwen2025qwenimage},  FLUX.2 [klein]~\cite{bfl2026flux2klein} and iMontage \cite{fu2025imontage} handle multi-image inputs but are not optimized for this setting and can be unstable when content preservation, style alignment, and instruction following must all be satisfied simultaneously.
Among task-specific methods, USO~\cite{wu2025uso} constructs compositional content-style triplets through subject-driven generation and de-stylization, representing an important step toward dual-reference control.
Nevertheless, the field still lacks large-scale, diverse, and cleanly separated triplet data, as well as systematic benchmarks evaluating the trade-offs among content fidelity, style alignment, instruction following, and leakage suppression.

\subsection{Data Construction for Controllable Generation}
High-quality triplet data consisting of content reference, style reference, and target image with clean separation, is fundamental to dual-reference generation, yet existing pipelines remain limited in scale, diversity, or separation quality.
Preference-based datasets such as Premier~\cite{wang2026premier} mine user interactions but are not designed for content-style disentanglement.
Synthetic stylization methods such as MegaStyle~\cite{gao2026megastyle} and OmniStyle~\cite{wang2025omnistyle} scale more easily but inherit the style range and artifacts of the underlying generator.
Hybrid pipelines such as TeleStyle~\cite{zhang2026telestyle} improve quality through manual curation yet are difficult to scale to broad style categories.
De-stylization-based methods such as USO~\cite{wu2025uso} directly target compositional supervision but may weaken structural details during content recovery.
In contrast, we mine community LoRAs as scalable compositional anchors for style and content, constructing large-scale triplets with broad style diversity and clean content-style separation across multiple base models.

\section{Method Overview}\label{sec:overview}

We present FreeStyle, a framework for style- and content-dual-reference image generation built on three tightly coupled components.

\paragraph{Data (\S\ref{sec:data}).}
We construct two complementary datasets.
A style-transfer pipeline (\S\ref{sec:style_data}) generates triplets by applying controlled stylization to diverse content images, providing clean supervision for basic style-reference generation.
A community-LoRA mining pipeline (\S\ref{sec:lora_pipeline}), the central data contribution of this work, treats LoRA weights as compositional anchors for both style and content, enabling large-scale dual-reference triplet construction with broad long-tail style coverage.

\paragraph{Training (\S\ref{sec:training}).}
We adopt a two-stage curriculum progressing from style-reference generation (Stage~1, trained on style-transfer data) to the harder dual-reference setting (Stage~2, mixing LoRA-mined triplets with style-transfer data).
Each stage faces a distinct content-leakage mechanism and employs a corresponding disentanglement strategy: an attention-level enrichment constraint (\S\ref{sec:attention_constraint}) for Stage~1 and frequency-aware RoPE modulation (\S\ref{sec:rope_modulation}) for Stage~2.

\paragraph{Benchmark (\S\ref{sec:benchmark}).}
We introduce an open benchmark for both style-reference and dual-reference generation. It adopts the style-invariant Content Alignment Score (CAS) from CSGO~\cite{xing2024csgo} for content evaluation and introduces a  VLM-based Verification Score for measuring style-transfer and content-preservation reliability under potential cross-reference leakage.

\section{Data Pipeline}\label{sec:data}

To support two-stage training, we construct two complementary datasets.
We first describe the style-transfer data that provides supervision for Stage~1 (\S\ref{sec:style_data}), and then present the community-LoRA-based mining pipeline that produces large-scale dual-reference triplets for Stage~2 (\S\ref{sec:lora_pipeline}).

\subsection{Style-Transfer Data Construction}\label{sec:style_data}

\begin{figure}[t]
\centering
\includegraphics[width=\linewidth]{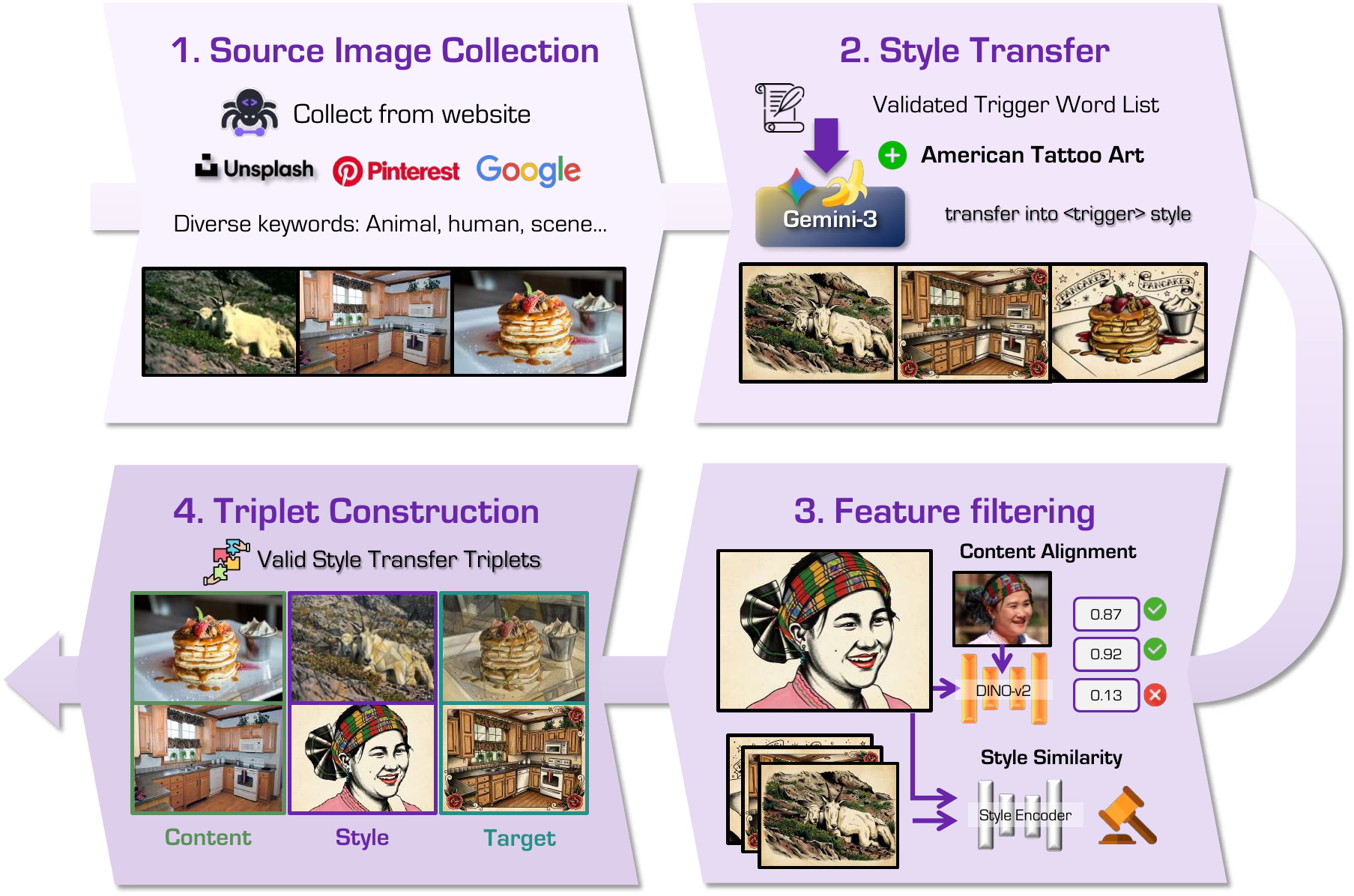}
\caption{\textbf{Overview of the Style-transfer Data Construction Pipeline.}
\textbf{(1)~Content collection.} We crawl a large set of raw content images from diverse websites, covering categories such as landscapes, human subjects, and everyday objects.
\textbf{(2)~Trigger-word stylization.} We validate a community style-trigger-word list and retain roughly 600 stable triggers, then stylize each content image through a fixed ``transfer into [trigger] style'' template.
\textbf{(3)~Bilateral consistency filtering.} For every stylized output we verify content fidelity against its source image via DINOv2 feature similarity, and measure style similarity with the ONEIG image encoder. Samples that pass both checks are assembled into clean style-transfer triplets.}
\label{fig:gemini_sref_pipeline}
\end{figure}

Stage~1 requires large-scale style-transfer triplets (content image, style reference, stylized target) with clean separation between content structure and visual style.
We construct this dataset by leveraging the state-of-the-art generative model Nano Banana Pro \cite{google2026nanobananapro} combined with bilateral consistency filtering, as illustrated in Figure~\ref{fig:gemini_sref_pipeline}.

Specifically, we first curate diverse content images from the web spanning landscapes, human subjects, and everyday objects, and validate approximately 645 style trigger words for generation stability.
Each content image is then stylized through a fixed prompt template (e.g., ``transfer into [trigger] style'').
We apply bilateral consistency filtering to every output: content fidelity is verified against the source via DINOv2~\cite{oquab2023dinov2} feature similarity, and style consistency is measured against anchor style images using the ONEIG style encoder.
Only samples passing both checks are retained.
When assembling triplets, the style reference is sampled from a different content source to ensure content-style independence, and text prompts are drawn from a GPT \cite{openai2023gpt4}-generated pool of style-transfer instructions.

\subsection{Community LoRA Mining and Triplet Construction}\label{sec:lora_pipeline}

Thanks to the continuous evolution of ComfyUI \cite{comfyanonymous2023comfyui}, we observe that the open-source community hosts a massive amount of high-quality LoRA models and creative workflows.
These LoRAs are exceptionally rich and diverse, covering an extensive range of content categories including characters, architecture, scenes, animals, daily objects, food, and vehicles, as well as nearly all possible stylistic categories.
However, constrained by factors such as the intrinsic instability of certain LoRA weights, the potential of these open-source resources has not been fully exploited.
To this end, we mine community LoRAs from platforms like Civitai \cite{civitai}, TensorArt \cite{tensorart}, and Liblib \cite{liblibai}, and design a highly robust pipeline for filtering and image generation, ultimately constructing a large-scale dataset for dual-reference generation.

\begin{figure*}[t]
\centering
\includegraphics[width=\textwidth,keepaspectratio]{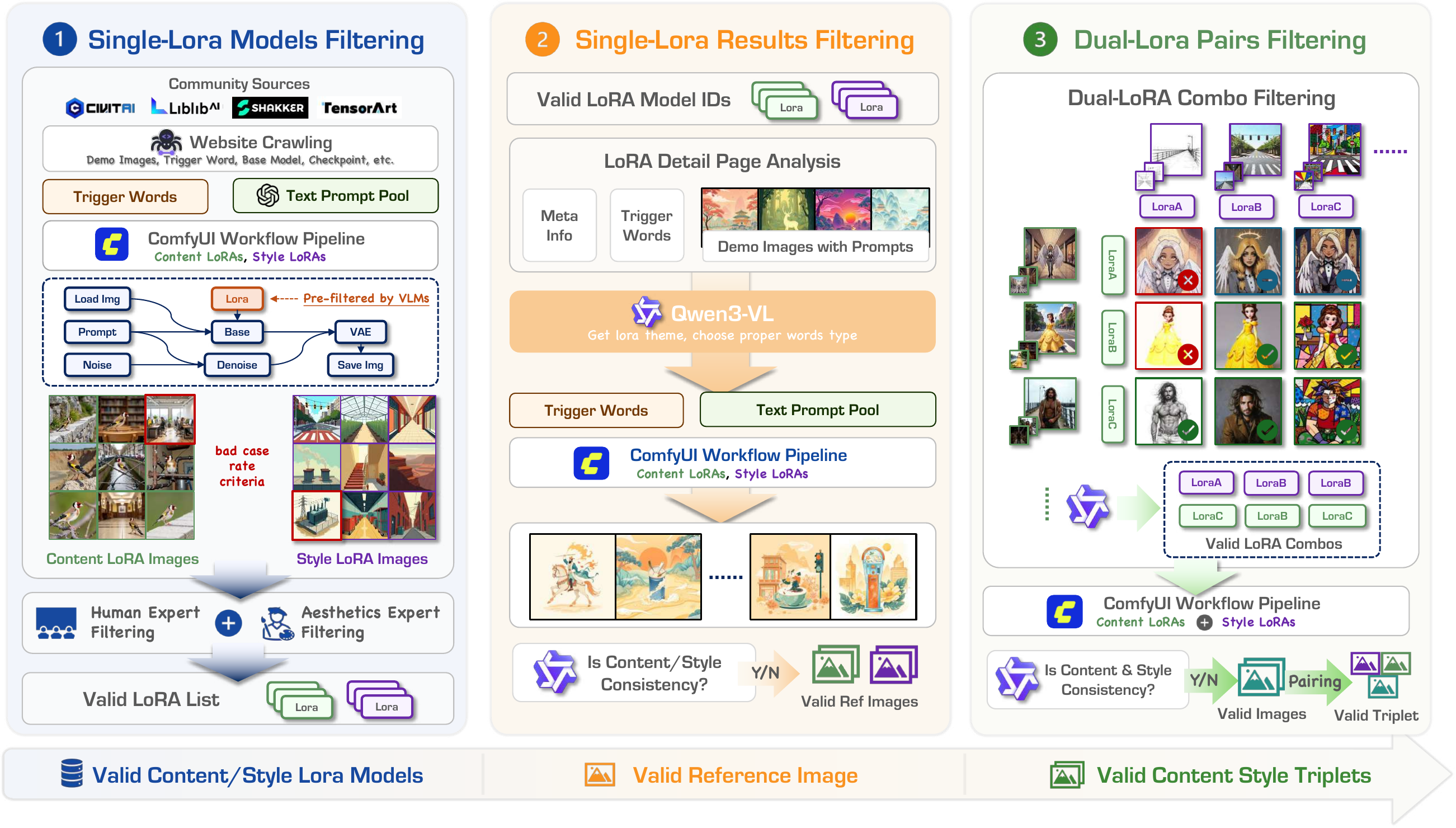}
\vspace{-8mm}
\caption{\small \textbf{Overview of the FreeStyle Data Construction Pipeline}.
\highlighttext{white}{pipelineblue}{\textbf{1}}~~\textbf{\textcolor{pipelineblue}{Single-LoRA Models Filtering}.} Community content and style LoRAs are collected and curated to build a high-quality LoRA repository.
\highlighttext{white}{pipelineorange}{\textbf{2}}~~\textbf{\textcolor{pipelineorange}{Single-LoRA Results Filtering}.} Representative reference images are identified through metadata analysis and generation-based validation.
\highlighttext{white}{pipelinegreen}{\textbf{3}}~~\textbf{\textcolor{pipelinegreen}{Dual-LoRA Pairs Filtering}.} Compatible content--style LoRA pairs are selected and combined to construct high-quality content--style triplets.}
\label{fig:sref_pipeline}
\end{figure*}

We first crawl the LoRA weights along with their web metadata.
Categorizing them by their base models, we select LoRA weights built upon three well-established text-to-image backbones: Illustrious \cite{park2024illustrious}, FLUX-dev \cite{flux2024}, and Qwen-Image \cite{qwen2025qwenimage}.
For each base model, we carefully design and tune specific ComfyUI workflows.
To fully unleash the effectiveness of these LoRAs while ensuring diversity, we draw on community-summarized heuristics to design vocabularies that are highly compatible with both style and content generation.
Furthermore, drawing inspiration from the category taxonomy of the OpenImages dataset \cite{kuznetsova2020openimage}, we incorporate extensive object-related vocabularies to enhance the content richness of the generated style reference images.
Based on these vocabularies, we sample and construct a massive prompt pool that allows for diverse random prompt generation.
With these preparations, we execute the following steps to obtain rich triplet images, as illustrated in the overall pipeline in Figure~\ref{fig:sref_pipeline}.

\begin{figure*}[tb]
\centering
\includegraphics[width=\textwidth]{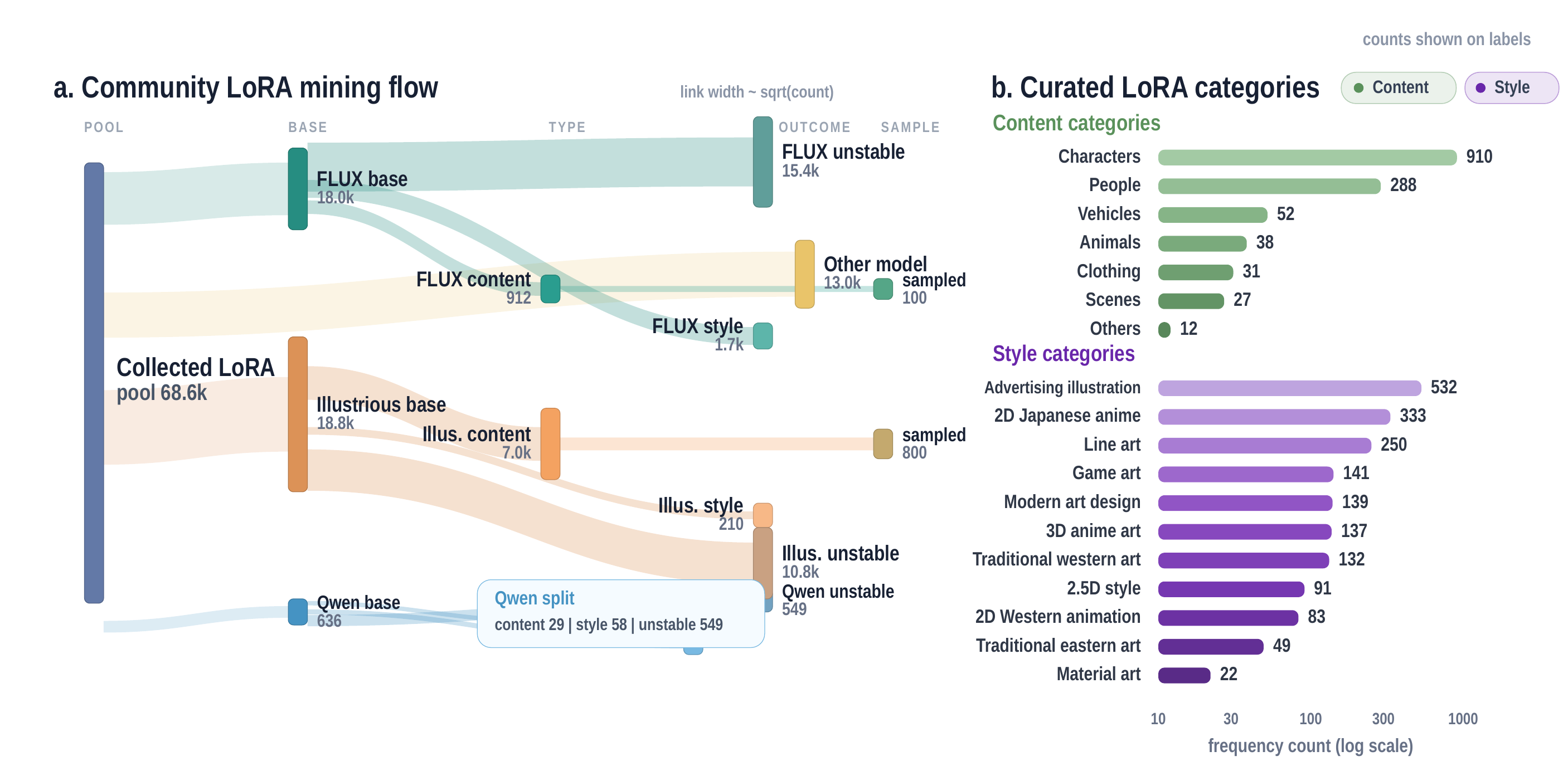}
\caption{\textbf{Statistics of Community LoRA Mining and Filtering.} The Sankey diagram (left) traces the successive stages of our LoRA-filtering pipeline: raw community LoRAs first undergo stability screening, and, to curb the combinatorial explosion of style--content pairings, we additionally sub-sample the content LoRAs via aesthetic-score-weighted non-uniform sampling, where higher-rated LoRAs are retained with higher probability, yielding the final set of curated weights. The distribution (right) classifies these final LoRAs by theme, following the same taxonomy as our soon-to-be-released benchmark.}
\label{fig:sankey_lora_mining}
\end{figure*}

\paragraph{1. Collection and Filtering of Stable LoRA Weights.}
To ensure a high success rate for subsequent combinations, we must first screen out reliable style and content LoRAs.
To make the generation compatible with both types of LoRAs, the prompt pool in this stage primarily consists of scene-related vocabulary combined with the inherent trigger words of each LoRA.
We generate a $3\times3$ grid of 9 preview images for human experts to evaluate whether the generation is stable and to categorize the LoRA as either content-oriented or style-oriented.
A LoRA is considered stably triggered only if at least 7 out of the 9 images in the grid exhibit consistent quality.
Once the stable content and style LoRAs are obtained, given their sheer volume, we perform aesthetic scoring \cite{verb2024aesthetic} and ranking on the content LoRAs to preserve content diversity and reduce the number of candidate combinations.
We then apply non-uniform sampling based on these scores: higher-scoring content LoRAs have a higher probability of being sampled, thereby reducing the effective scale of the LoRA pool.
Figure~\ref{fig:sankey_lora_mining} illustrates the Sankey diagram of our LoRA filtering process and the final distribution of curated LoRAs, where the style categorization aligns with the taxonomy used in our benchmark.

\paragraph{2. Generation of Rich and Effective Reference Images.}
With the stable LoRAs secured, we use them to generate reference images that are both content-rich and highly representative of the LoRAs' own themes.
Here, we utilize the comprehensive prompt pool prepared initially, which contains approximately 40k distinct prompt combinations, and combine them with the LoRAs' trigger words.
At least 20 images are generated for each LoRA.
Since the generated outputs are not always perfectly stable, we use the preview images collected from the web as references and employ Qwen3-VL \cite{qwen3vl} for verification.
The verification adopts the binary VLM judgment and majority-voting protocol used in our benchmark, yielding accurate and stable filtering results.
Notably, we design distinct parameters and workflows for different base models to perform batch generation via ComfyUI services; these workflows will also be open-sourced.

\paragraph{3. Filtering of Valid LoRA Combinations.}
LoRAs can be combined with one another.
In many excellent community workflows, multiple LoRAs are often applied simultaneously to a single base model using various advanced techniques (e.g., community experience suggests that inserting LoRAs into different layers of SDXL \cite{podell2024sdxl} yields diverse effects).
Although style and content LoRAs do not inherently conflict in their domains, the combined effect largely depends on the stability of the LoRAs themselves.
However, we noticed that the content of some LoRAs carries intrinsic stylistic traits, leading to degradation and instability during actual combination.
In practice, this instability is persistent: if a combination has inherent conflicts, batch testing with even highly diverse prompts will still yield poor results.
Therefore, LoRA combinations must be rigorously filtered.
To address this, we perform a Cartesian product of the style and content LoRAs, generate one image per combination, and conduct bilateral content and style verification against the single-LoRA images generated previously.
In this step, the individual filtering success rate is approximately 0.6 for style and 0.8 for content, while the success rate for the bilateral combination drops to about 0.4.
Through this rigorous process, we acquire a massive set of stable and reliable LoRA combinations.

\paragraph{4. Batch Generation of Target Triplet Data.}
Having established stable LoRA combinations, we leverage the rich vocabulary data to batch-generate massive target images for our triplets.
These target images are again subjected to bilateral consistency verification against the corresponding single-LoRA style and content reference images.
Importantly, because the validity of the LoRA combinations has already been screened, the verification success rate during batch generation reaches as high as 0.8, dramatically boosting efficiency and stability.
Ultimately, we successfully generated and filtered 273k triplets using the FLUX model, 33k using the Qwen model, and 172k using the Illustrious model.
The resulting dataset is exceptionally massive and highly diverse in its compositional combinations.

Finally, we employ a Qwen3-VL model \cite{qwen3vl} to generate the image content prompts that guide the model in bridging the content reference image and the target image.
Through these meticulous steps, we assemble a comprehensive and highly diverse triplet dataset consisting of the content reference image, the style reference image, the generative prompt, and the target image.

\section{Style-Content Disentanglement Training}\label{sec:training}

Our method introduces no additional architectural modifications to the base model.
We inject text, content-reference, and style-reference conditions into the MMDiT blocks~\cite{vaswani2017attention,dosovitskiy2021vit,peebles2023dit,esser2024sd3}, following the multi-image input paradigm of Qwen-Image-Edit, whose text branch is encoded by a vision-language model~\cite{bai2025qwen25vl} and whose generator is trained with a flow-matching objective~\cite{lipman2023flowmatching,liu2023rectifiedflow}.

We adopt a two-stage training curriculum.
In \textbf{Stage~1}, the model is trained exclusively on style-transfer data (Sec.~\ref{sec:style_data}) to build robust style-reference generation capability.
At this stage, content leakage---where semantic elements from the style reference are inadvertently copied into the generated image---primarily manifests as disproportionate attention allocation to style-reference tokens during late denoising steps.
We address this with an attention-level enrichment constraint (Sec.~\ref{sec:attention_constraint}).
In \textbf{Stage~2}, we mix community-LoRA-mined dual-reference triplets (Sec.~\ref{sec:lora_pipeline}) with style-transfer data, introducing the harder setting where the model must simultaneously respect a content reference, a style reference, and a text instruction.
Now, the content reference absorbs a substantial share of the model's attention, and the style-reference attention no longer exhibits the same pronounced asymmetry seen in Stage~1.
Yet content leakage persists through a different channel: local positional correspondence encoded by high-frequency RoPE components, which enables patch-level copying from the style image.
We address this with frequency-aware RoPE modulation (Sec.~\ref{sec:rope_modulation}).
The two stages thus target distinct leakage mechanisms and employ complementary disentanglement strategies.

\subsection{Attention Constraint for Style-Reference Generation}\label{sec:attention_constraint}

\begin{figure*}[t]
\centering
\includegraphics[width=\textwidth]{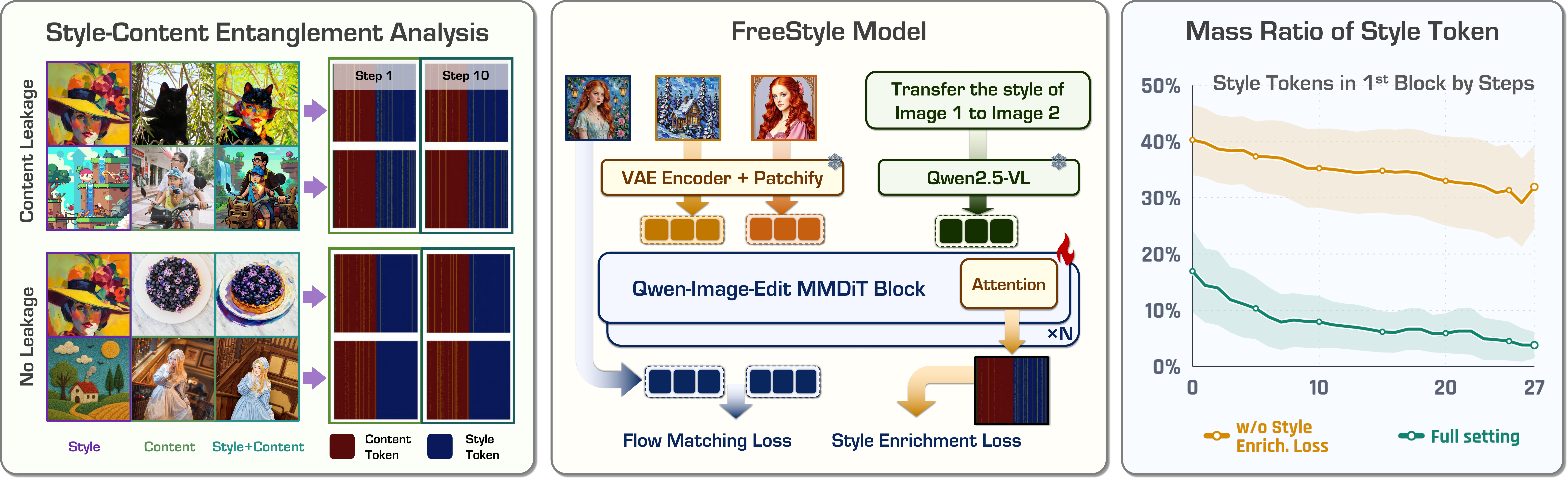}
\caption{\textbf{Analyzing Content Leakage through the Lens of Attention.}
\textbf{Left:} failure cases exhibiting semantic content leakage from the style reference, contrasted with successful leakage-free transfers, where leakage cases produce broader and more persistent high-response bands than successful transfers. This observation motivates our group-wise attention-enrichment constraint.
\textbf{Middle:} our dual-reference architecture together with the style-reference attention mass-ratio dynamics across denoising time in the first transformer block.
\textbf{Right:} Attention mass ratio evaluation across different time steps in first block, the orange curve denotes the variant without the style-enrichment loss, whereas the green curve denotes the full model.}
\label{fig:attention_enrichment_combined}
\end{figure*}

\paragraph{Observation: content leakage correlates with attention asymmetry.}
Before designing any constraint, we first analyze how attention allocation over different semantic groups correlates with content leakage in style-reference generation.
Figure~\ref{fig:attention_enrichment_combined} presents representative evidence.
In the attention maps (left panel), the vertical axis corresponds to query-side noisy-latent tokens and the horizontal axis to semantic partitions on the key side, with the dark-red and dark-blue regions indicating content-reference and style-reference token ranges, respectively.
Comparing leakage cases against successful transfers reveals a clear pattern: \emph{leakage cases exhibit substantially broader and more persistent high-response bands over the style-reference region}, whereas successful cases maintain more compact and stable responses.
The right panel further shows the style-reference attention mass ratio across denoising time in first transformer block, confirming that the asymmetry intensifies in late denoising steps and is most prominent at the first transformer block, where the global semantic layout of the generated image is determined.

This observation reveals that content leakage is not a static model property, but a \emph{time-varying failure pattern} that can be localized along both the denoising and depth axes.
It motivates us to (i) define a quantitative metric that captures this disproportionate allocation, and (ii) convert it into a differentiable constraint.

\paragraph{Group-wise attention enrichment.}
To quantify the above phenomenon, we measure how much attention each semantic group of tokens receives relative to its size.
Let $A^{(t,\ell)}$ be the attention map at denoising step $t$ and transformer block $\ell$, with entry $A_{qk}$ the attention weight from a noisy-latent query $q$ to a key token $k$.
We split the keys into semantic groups---text, content reference (\texttt{cref}), and style reference (\texttt{sref})---writing $\mathcal{G}_g$ for group $g$ and $\mathcal{K}$ for all keys.
The share of attention that group $g$ receives is
\begin{equation}
p_g^{(t,\ell)}
=
\frac{
\sum_{q}\sum_{k\in\mathcal{G}_g} A_{qk}^{(t,\ell)}
}{
\sum_{q}\sum_{k\in\mathcal{K}} A_{qk}^{(t,\ell)}
},
\end{equation}
A larger group naturally attracts more attention, so we normalize by the group's size fraction to obtain the \emph{enrichment score}:
\begin{equation}
E_g^{(t,\ell)}=\frac{p_g^{(t,\ell)}}{|\mathcal{G}_g|/|\mathcal{K}|}.
\label{eq:enrich_score}
\end{equation}
Here $E_g=1$ means group $g$ is attended to exactly in proportion to its size; $E_g>1$ indicates enrichment (the group draws more attention than its size warrants); and $E_g<1$ indicates suppression.
Because the score factors out group size, it places groups of different sizes on equal footing while still resolving how attention shifts along both the denoising and depth axes.

\paragraph{Attention-regularization losses.}
Based on the above analysis, we convert the enrichment metric into two lightweight regularizers.
Both are applied only at the first transformer block ($\ell=0$), where attention over the style reference is most predictive of leakage.
The two losses share a single two-sided squared hinge that keeps a quantity $x$ within a target band $[a,b]$:
\begin{equation}
\psi_{[a,b]}(x) = \max(0,\,a-x)^2 + \max(0,\,x-b)^2 .
\end{equation}

\emph{Enrichment loss.}
For each noisy-latent query $q$, the style-reference enrichment at the first block follows Eq.~\eqref{eq:enrich_score} restricted to that query, $E_{\mathrm{sref}}(t,q)=\bigl(\textstyle\sum_{k\in\mathcal{G}_{\mathrm{sref}}}A_{qk}^{(t,1)}\bigr)\big/\bigl(|\mathcal{G}_{\mathrm{sref}}|/|\mathcal{K}|\bigr)$.
We keep it inside $[\alpha_{\mathrm{lo}},\alpha_{\mathrm{hi}}]$ and weight later denoising steps more heavily by $(1-t)$:
\begin{equation}
\mathcal{L}_{\mathrm{enrich}} = \mathbb{E}_{t,\,q}\Bigl[(1-t)\,\psi_{[\alpha_{\mathrm{lo}},\,\alpha_{\mathrm{hi}}]}\bigl(E_{\mathrm{sref}}(t,q)\bigr)\Bigr].
\end{equation}
We set $\alpha_{\mathrm{lo}}=0$ and $\alpha_{\mathrm{hi}}=0.6$: the zero lower bound means we never force attention onto the style reference and only penalize over-attention beyond $0.6\times$ the size-matched baseline, which is what drives content copying.
The $(1-t)$ weight concentrates the constraint on late denoising steps, where reference influence actually materializes.

\emph{Entropy loss.}
Suppressing over-attention can collapse the style-reference attention onto a few keys and weaken style transfer.
We therefore also regularize the shape of each query's distribution $\tilde{A}_{qk}=A_{qk}^{(t,1)}/\sum_{k'\in\mathcal{G}_{\mathrm{sref}}}A_{qk'}^{(t,1)}$ over $\mathcal{G}_{\mathrm{sref}}$ through its normalized entropy
\begin{equation} \hat{\mathcal{H}}_q = -\frac{1}{\log|\mathcal{G}_{\mathrm{sref}}|}\sum_{k\in\mathcal{G}_{\mathrm{sref}}}\tilde{A}_{qk}\log\tilde{A}_{qk}\;\in[0,1], \end{equation}
keeping it inside a narrow band with the same hinge:
\begin{equation}
\mathcal{L}_{\mathrm{ent}} = \mathbb{E}_{t,\,q}\Bigl[\psi_{[\beta_{\mathrm{lo}},\,\beta_{\mathrm{hi}}]}\bigl(\hat{\mathcal{H}}_q\bigr)\Bigr].
\end{equation}
We use $\beta_{\mathrm{lo}}=0.06$ and $\beta_{\mathrm{hi}}=0.14$: below the band the attention collapses onto a few style tokens, above it the attention smears out uniformly and loses focus.

The full objective augments the flow-matching diffusion loss $\mathcal{L}_{\mathrm{diff}}$ with both regularizers:
\begin{equation}
\mathcal{L} = \mathcal{L}_{\mathrm{diff}} + \lambda_{\mathrm{e}}\,\mathcal{L}_{\mathrm{enrich}} + \lambda_{\mathrm{h}}\,\mathcal{L}_{\mathrm{ent}},
\end{equation}
where $\lambda_{\mathrm{e}}=\lambda_{\mathrm{h}}=0.1$ so the two terms gently shape attention without overpowering the main objective.
Together they keep content leakage under control while preserving style diversity.

\subsection{Frequency-Aware RoPE Modulation for Dual-Reference Generation}\label{sec:rope_modulation}

After Stage~1, the model has acquired robust style-reference generation with content leakage well controlled by the attention constraint.
However, when we transition to the dual-reference setting in Stage~2, the nature of the task fundamentally changes: the model must now attend to a content reference, a style reference, and a text prompt simultaneously, using the content reference as the structural scaffold while transferring style from the style reference.
In this joint-conditioning regime, the content reference absorbs a large share of the model's attention budget, and the style-reference enrichment score no longer exhibits the pronounced late-stage spike observed in the style-only setting.
As a result, the attention-based enrichment constraint from Sec.~\ref{sec:attention_constraint} loses much of its discriminative power.

Yet content leakage from the style reference persists---but through a different mechanism.
We hypothesize that the leakage pathway shifts from \emph{attention magnitude} to \emph{positional correspondence}: the high-frequency components of RoPE~\cite{su2024roformer} encode fine-grained spatial relationships between tokens, which can enable the model to establish patch-level copying mappings between the style reference and the generated output.
Inspired by the frequency-aware modulation strategy in~\cite{mikaeili2026untwistingrope}, we regularize the rotary positional embeddings of the style-reference branch.
The core idea is to suppress high-frequency RoPE components that encourage local copying, while amplifying low-frequency components that better preserve global stylistic structure.

Concretely, let $d \in \{0,\ldots,D/2-1\}$ index the two-dimensional RoPE chunks and let $D$ denote the RoPE dimensionality.
We assign a smooth frequency-dependent scale to the \texttt{sref} keys:
\begin{equation}
s_d
=
s_{\mathrm{hf}}
+ \Bigl(s_{\mathrm{lf}}-s_{\mathrm{hf}}\Bigr)
\left(\frac{d}{D/2-1}\right)^{\beta},
\end{equation}
where $s_{\mathrm{hf}} < 1$ suppresses high-frequency locality, $s_{\mathrm{lf}} > 1$ amplifies low-frequency global guidance, and $\beta$ controls the smoothness of the interpolation across frequency bands. These scaling factors are fixed throughout the denoising process and do not depend on the timestep. We apply this modulation only to the style-reference branch, while leaving the content-reference branch unchanged.

\section{Benchmark}\label{sec:benchmark}

Existing evaluations of style-reference and dual-reference generation are fragmented: different works report different metrics on private test sets, making cross-method comparison unreliable.
To address this, we introduce an open benchmark with a fixed reference set, standardized prompts, and a mixed evaluation protocol that separates style fidelity, content preservation, and leakage control into distinct axes rather than collapsing them into a single score.
We will publicly release the benchmark, prompts, and evaluation code.

\subsection{Benchmark Construction}\label{sec:bench_construction}

The benchmark is built from \textbf{200 content reference images} and \textbf{200 style reference images}, curated to maximize diversity.
Content images span a broad range of object categories, scenes, and compositions, while style images cover artistic domains from oil painting and watercolor to 3D rendering, pixel art, and abstract art.
Reference pairings are sampled \emph{without replacement} to prevent any single style or content from dominating the evaluation.

For each pairing, the text prompt is drawn from a GPT-generated \cite{openai2023gpt4} prompt pool of style-transfer-oriented instructions.
We generate multiple prompt variants per style category to reduce prompt-specific bias.
The benchmark covers two evaluation settings:
\begin{itemize}
    \item \textbf{Style-reference generation (SRef):} given a source image, a style-reference image, and a text instruction, generate an image that preserves the source content while adopting the visual style of the style reference.
    \item \textbf{Content-style dual-reference generation (CRef+SRef):} given a content reference image, a style reference image, and a text prompt, generate an image that preserves the structure of the content reference, transfers the style of the style reference, and respects the text instruction.
\end{itemize}

\subsection{Evaluation Metrics}\label{sec:eval_metrics}

We adopt a mixed evaluation protocol combining feature-based encoder similarities, VLM-based judgments, and aesthetic scores.
Each evaluation axis (style fidelity, content fidelity, instruction following) includes both an encoder-based metric and a VLM-based metric, since the two capture complementary aspects: encoder similarities measure continuous perceptual distance, while VLM judgments assess higher-level semantic alignment that single-axis feature distances may not fully reflect.
All VLM-based scores (VLM-S, VLM-C, VLM-F, Ver-S, Ver-C) are computed using Qwen3-VL~\cite{qwen3vl} as the judge.

\paragraph{Style-fidelity axis.}
\textbf{ONEIG} \cite{chang2026oneig} provides a perceptual estimate of stylistic consistency.
\textbf{CSD}~\cite{somepalli2024csd} measures style similarity using a contrastive descriptor trained to separate style from content.
\textbf{VLM-S} (VLM-Style) asks a vision-language model to rate how well the generated image adopts the overall visual style of the reference, yielding a scalar score.

\paragraph{Content-fidelity axis.}
\textbf{DINOv2}~\cite{oquab2023dinov2} reports cosine similarity between the generated image and the content reference in the self-supervised DINOv2 embedding space, sensitive to object layout and semantic structure.
\textbf{CAS} (Content Alignment Score), adopted from CSGO~\cite{xing2024csgo} and detailed below, measures structural agreement after removing style-carrying channel statistics, thereby reducing the influence of appearance differences on content evaluation.
\textbf{VLM-C} (VLM-Content) asks a vision-language model to rate how well the generated image preserves the structure and semantics of the content reference.

\paragraph{Instruction-following axis (CRef+SRef only).}
\textbf{CLIP-T}~\cite{radford2021clip} reports image-text cosine similarity in CLIP space.
\textbf{VLM-F} (VLM-Follow) asks a vision-language model to judge whether the generated image obeys the text instruction, yielding a scalar score.

\paragraph{VLM verification.}
Ver-S and Ver-C (defined below) measure the proportions of generated images that are verified as successful in style transfer and content preservation, respectively, using repeated binary VLM judgments with majority voting.

\paragraph{Aesthetics.}
\textbf{LAION-Aesthetic}~\cite{laion_aesthetic_predictor} and \textbf{V2.5-Aesthetic}  score \cite{verb2024aesthetic} the raw visual appeal of each output, guarding against degenerate solutions that satisfy similarity metrics but look unnatural.

\paragraph{Content Alignment Score (CAS).}
We adopt the Content Alignment Score proposed in CSGO~\cite{xing2024csgo} to evaluate content preservation while reducing sensitivity to stylistic appearance. CAS builds on the observation that channel-wise feature statistics capture appearance-related information, whereas instance-normalized features retain more of the underlying spatial structure~\cite{huang2017adain}. Specifically,
let $\phi(\cdot)\in\mathbb{R}^{L\times C}$ denote the DINOv2~\cite{oquab2023dinov2} patch-token features of an image resized to $512\times512$, where $L$ is the number of tokens and $C$ the channel dimension.
For each channel $c$, we compute its mean and standard deviation over the $L$ tokens,
\begin{equation}
\begin{aligned}
\mu_c(I)&=\frac{1}{L}\sum_{l=1}^{L}\phi(I)_{l,c},\\
\sigma_c(I)&=\sqrt{\frac{1}{L}\sum_{l=1}^{L}\bigl(\phi(I)_{l,c}-\mu_c(I)\bigr)^2+\epsilon},
\end{aligned}
\end{equation}
and apply instance normalization to remove the style-carrying statistics:
\begin{equation}
\hat{\phi}(I)_{l,c}=\frac{\phi(I)_{l,c}-\mu_c(I)}{\sigma_c(I)}.
\end{equation}
Given a generated image $I_g$ and its content reference $I_c$, CAS is the mean squared error between their style-normalized features:
\begin{equation}
\mathrm{CAS}=\frac{1}{LC}\sum_{l=1}^{L}\sum_{c=1}^{C}\bigl(\hat{\phi}(I_g)_{l,c}-\hat{\phi}(I_c)_{l,c}\bigr)^2 .
\end{equation}
A lower CAS indicates better content preservation, since the metric isolates structural agreement after factoring out style differences.

\paragraph{Verification Score.}
Feature-based metrics measure average similarity but do not directly indicate whether individual generations reliably satisfy the desired style-transfer and content-preservation criteria.
To complement them, we introduce a VLM-based Verification Score that aggregates repeated binary judgments.

Given the token log-probabilities $\ell_0$ and $\ell_1$ assigned by the VLM to the outputs  ``0'' and ``1'', respectively, we normalize them over the two candidate labels:
\begin{equation}
p_c =
\frac{\exp(\ell_c)}
{\exp(\ell_0) + \exp(\ell_1)},
\qquad c \in \{0,1\}.
\end{equation}
The predicted binary label is
\begin{equation}
\hat{y} =
\operatorname*{arg\,max}_{c \in \{0,1\}} p_c.
\end{equation}
We query the VLM three times for each generated--reference image pair.
A pair is counted as a successful match if at least two of the three judgments predict $\hat{y}=1$.
Because each generated image is evaluated along two independent axes, we report two Verification Scores:
\begin{itemize}
\item \textbf{Style Verification Score (Ver-S):} measures the proportion of generated images judged to correctly transfer the target style.
\item \textbf{Content Verification Score (Ver-C):} measures the proportion of generated images judged to preserve the content reference under potential semantic leakage from the style image.
\end{itemize}
Higher values indicate more reliable style transfer and content preservation.
The Verification Score complements feature-based similarities by providing a binary pass/fail signal, offering an additional perspective on whether the model successfully balances style transfer and content preservation.



\begin{table*}[t]
\centering
\caption{\textbf{Quantitative Comparison on the \textit{Style-reference (SRef)} Benchmark}.
Metrics are grouped by evaluation axis: style fidelity (feature-based and VLM-based), content fidelity, VLM-based verification, and aesthetics.
Closed-source models are shown above the rule for reference only.
Among \emph{open-source methods}, the three purple shades mark the top-3 results per metric, with our model at the bottom.
$\uparrow$: higher is better; $\downarrow$: lower is better.}
\label{tab:sref_benchmark}
\vspace{-2mm}
\scriptsize
\setlength{\tabcolsep}{4.5pt}
\renewcommand{\arraystretch}{1.2}
\begin{tabular}{@{}l*{10}{c}@{}}
\toprule
\multirow{2}{*}{\textbf{Method}}
& \multicolumn{3}{c}{\textbf{Style}}
& \multicolumn{3}{c}{\textbf{Content}}
& \multicolumn{2}{c}{\textbf{VLM Verification}}
& \multicolumn{2}{c}{\textbf{Aesthetics}} \tabularnewline
\cmidrule(lr){2-4}
\cmidrule(lr){5-7}
\cmidrule(lr){8-9}
\cmidrule(lr){10-11}
& ONEIG$\uparrow$
& CSD$\uparrow$
& VLM-S$\uparrow$
& DINO$\uparrow$
& CAS$\downarrow$
& VLM-C$\uparrow$
& Ver-S$\uparrow$
& Ver-C$\uparrow$
& LAION$\uparrow$
& V2.5$\uparrow$ \tabularnewline
\midrule
\multicolumn{11}{@{}l}{\textit{Closed-source commercial models}} \tabularnewline
Gemini       & 0.416 & 0.619 & 7.848 & 0.840 & 1.452 & 8.798 & 0.554 & 0.852 & 6.157 & 5.749 \tabularnewline
Seedream     & 0.441 & 0.636 & 7.023 & 0.836 & 1.077 & 9.154 & 0.334 & 0.905 & 6.495 & 5.923 \tabularnewline
\midrule
\multicolumn{11}{@{}l}{\textit{Open-source methods}} \tabularnewline
USO          & \rkone{0.542} & 0.531 & 3.744 & 0.808 & 1.282 & \rktwo{9.237} & \rkthree{0.382} & \rktwo{0.928} & 5.971 & 5.575 \tabularnewline
CSGO         & \rktwo{0.520} & \rkone{0.665} & \rkthree{6.193} & 0.652 & 1.656 & 1.516 & 0.038 & 0.715 & 5.430 & 4.676 \tabularnewline
EasyRef      & 0.270 & 0.578 & 2.073 & 0.612 & 1.904 & 0.428 & 0.000 & 0.129 & 5.403 & 4.415 \tabularnewline
FLUX.2 [klein]   & \rkthree{0.490} & \rktwo{0.655} & 6.068 & \rkthree{0.811} & \rkthree{1.084} & 8.335 & 0.285 & 0.532 & \rkone{6.682} & \rkone{6.018} \tabularnewline
TeleStyle    & 0.459 & 0.613 & 5.555 & \rktwo{0.859} & \rktwo{0.924} & \rkone{9.566} & \rktwo{0.449} & \rkone{0.961} & \rkthree{6.173} & 5.470 \tabularnewline
Qwen-Image-Edit    & 0.290 & 0.589 & 3.487 & \rkone{0.865} & \rkone{0.918} & \rkthree{9.155} & 0.261 & \rkthree{0.910} & 6.131 & \rkthree{5.657} \tabularnewline
OmniStyle    & 0.389 & 0.603 & \rktwo{6.247} & 0.757 & 1.759 & 6.247 & 0.329 & 0.483 & 4.901 & 4.542 \tabularnewline
\textbf{Ours} & 0.468 & \rkthree{0.639} & \rkone{7.142} & 0.809 & 1.175 & 8.919 & \rkone{0.482} & \rktwo{0.928} & \rktwo{6.302} & \rktwo{5.709} \tabularnewline
\bottomrule
\end{tabular}
\end{table*}

\begin{table*}[t]
\centering
\caption{\textbf{Quantitative Comparison on the \textit{Dual-reference (CRef+SRef)} Benchmark}.
Metrics are grouped by evaluation axis.
Instruction-following metrics (CLIP-T and VLM-F) are grouped together since both measure prompt adherence.
Notation and shading conventions follow Table~\ref{tab:sref_benchmark}.}
\label{tab:dualref_benchmark}
\vspace{-2mm}
\scriptsize
\setlength{\tabcolsep}{3.5pt}
\renewcommand{\arraystretch}{1.2}
\begin{tabular}{@{}l*{12}{c}@{}}
\toprule
\multirow{2}{*}{\textbf{Method}}
& \multicolumn{3}{c}{\textbf{Style}}
& \multicolumn{3}{c}{\textbf{Content}}
& \multicolumn{2}{c}{\textbf{Instruction}}
& \multicolumn{2}{c}{\textbf{VLM Verification}}
& \multicolumn{2}{c}{\textbf{Aesthetics}} \tabularnewline
\cmidrule(lr){2-4}
\cmidrule(lr){5-7}
\cmidrule(lr){8-9}
\cmidrule(lr){10-11}
\cmidrule(lr){12-13}
& ONEIG$\uparrow$
& CSD$\uparrow$
& VLM-S$\uparrow$
& DINO$\uparrow$
& CAS$\downarrow$
& VLM-C$\uparrow$
& CLIP-T$\uparrow$
& VLM-F$\uparrow$
& Ver-S$\uparrow$
& Ver-C$\uparrow$
& LAION$\uparrow$
& V2.5$\uparrow$ \tabularnewline
\midrule
\multicolumn{13}{@{}l}{\textit{Closed-source commercial models}} \tabularnewline
Gemini       & 0.369 & 0.590 & 5.113 & 0.749 & 1.723 & 7.447 & 0.314 & 9.492 & 0.479 & 0.411 & 6.740 & 6.135 \tabularnewline
Seedream     & 0.372 & 0.601 & 5.621 & 0.759 & 1.607 & 7.723 & 0.319 & 9.565 & 0.522 & 0.477 & 6.969 & 5.593 \tabularnewline
\midrule
\multicolumn{13}{@{}l}{\textit{Open-source methods}} \tabularnewline
USO          & \rkone{0.450} & 0.246 & \rkthree{3.753} & \rkone{0.809} & \rkone{1.283} & \rkone{9.139} & 0.245 & 2.825 & \rktwo{0.390} & \rkone{0.916} & 5.943 & 5.566 \tabularnewline
Qwen-Image-Edit    & 0.232 & 0.489 & 2.386 & 0.723 & 1.731 & 6.141 & \rktwo{0.320} & \rktwo{9.220} & 0.118 & 0.320 & 6.587 & \rkthree{5.699} \tabularnewline
FLUX.2 [klein]   & 0.268 & \rkone{0.602} & 2.388 & \rkthree{0.758} & 1.702 & \rkthree{7.034} & \rkone{0.327} & \rkone{9.432} & 0.124 & 0.361 & \rkone{6.833} & \rktwo{5.734} \tabularnewline
TeleStyle    & \rkthree{0.362} & \rktwo{0.585} & \rktwo{4.251} & \rktwo{0.760} & \rktwo{1.590} & \rktwo{7.338} & 0.302 & 7.994 & \rkthree{0.315} & \rktwo{0.550} & \rktwo{6.800} & \rkone{5.774} \tabularnewline
\textbf{Ours} & \rktwo{0.387} & \rkthree{0.575} & \rkone{5.467} & 0.739 & \rkthree{1.639} & \rkthree{7.038} & \rkthree{0.308} & \rkthree{8.909} & \rkone{0.409} & \rkthree{0.462} & \rkthree{6.747} & 5.643 \tabularnewline
\bottomrule
\end{tabular}
\end{table*}

\section{Experiments}

We evaluate FreeStyle on the benchmark introduced in Sec.~\ref{sec:benchmark} under both the style-reference (SRef) and dual-reference (CRef+SRef) settings.
All metrics are defined in Sec.~\ref{sec:eval_metrics}; $\uparrow$ denotes higher-is-better and $\downarrow$ denotes lower-is-better.
For closed-source commercial baselines, we report the latest publicly available versions at the time of evaluation; in particular, the Seedream \cite{seedream2025seedream} entries correspond to \textbf{Seedream~4.5}, queried through its official API under the same reference images and prompts as all other methods.

\subsection{Implementation Details}\label{sec:impl_details}

We use the Qwen-Image-Edit-2511 \cite{qwen2025qwenimage} backbone and train with a flow-matching objective.
Stage~1 is trained on style-transfer data for 120k steps with batch size 32 and learning rate 6.0e-06.
Stage~2 mixes community-LoRA-mined triplets with style-transfer data (ratio 3:1) and trains for an additional 24k steps.
Both stages use 8$\times$ H100 GPUs.
The attention-enrichment constraint is applied from the beginning of Stage~1 with $\lambda_{\mathrm{e}}=\lambda_{\mathrm{h}}=0.1$.
Frequency-aware RoPE modulation is introduced at the start of Stage~2 with $\beta=2$, $s_{\mathrm{hf}}$ is 0.9, and $s_{\mathrm{lf}}$ is 1.2, following the experimental setup in \cite{mikaeili2026untwistingrope}.

\subsection{Benchmark Results}\label{sec:bench_results}

Tables~\ref{tab:sref_benchmark} and~\ref{tab:dualref_benchmark} report the SRef and CRef+SRef results, respectively.
We highlight several observations.

\paragraph{Feature-based and VLM-based metrics should be evaluated jointly.}
A natural trade-off exists between content-oriented and style-oriented metrics in this task. Models with conservative stylization may obtain high DINOv2 and CAS scores while achieving limited style transfer, whereas models with stronger stylization may obtain high ONEIG or CSD scores at the cost of content preservation or increased reference leakage.
For example, CSGO achieves the highest CSD score (0.665), while its relatively low VLM-Content (1.516) and Ver-C (0.715) suggest that the high feature-based style similarity does not necessarily correspond to reliable content preservation.
Conversely, Qwen-Image-Edit performs best on DINOv2 (0.865) and CAS (0.918), but its lower VLM-Style score (3.487) suggests comparatively limited style transfer.
These results illustrate that feature-based metrics alone do not fully distinguish successful style transfer from imbalanced content-style trade-offs; the feature-based metrics on each axis should therefore be read together with the corresponding VLM judgment and Verification Score.

\paragraph{FreeStyle achieves the best overall balance.}
On the SRef benchmark, our method ranks first on VLM-Style (7.142) and Ver-S (0.482), indicating the strongest VLM-verified style transfer among open-source methods, while maintaining competitive content preservation (Ver-C second at 0.928) and aesthetics (LAION second, V2.5 second).
On the harder CRef+SRef benchmark, the same pattern holds: FreeStyle ranks first on VLM-Style (5.467) and Ver-S (0.409), with competitive performance across all other axes.
No single baseline achieves comparable balance: USO excels on content metrics but scores poorly on style and instruction following; FLUX.2 [klein] leads on instruction metrics but transfers style weakly (VLM-Style 2.388).

\begin{figure*}[htbp]
\centering
\includegraphics[width=0.8\linewidth]{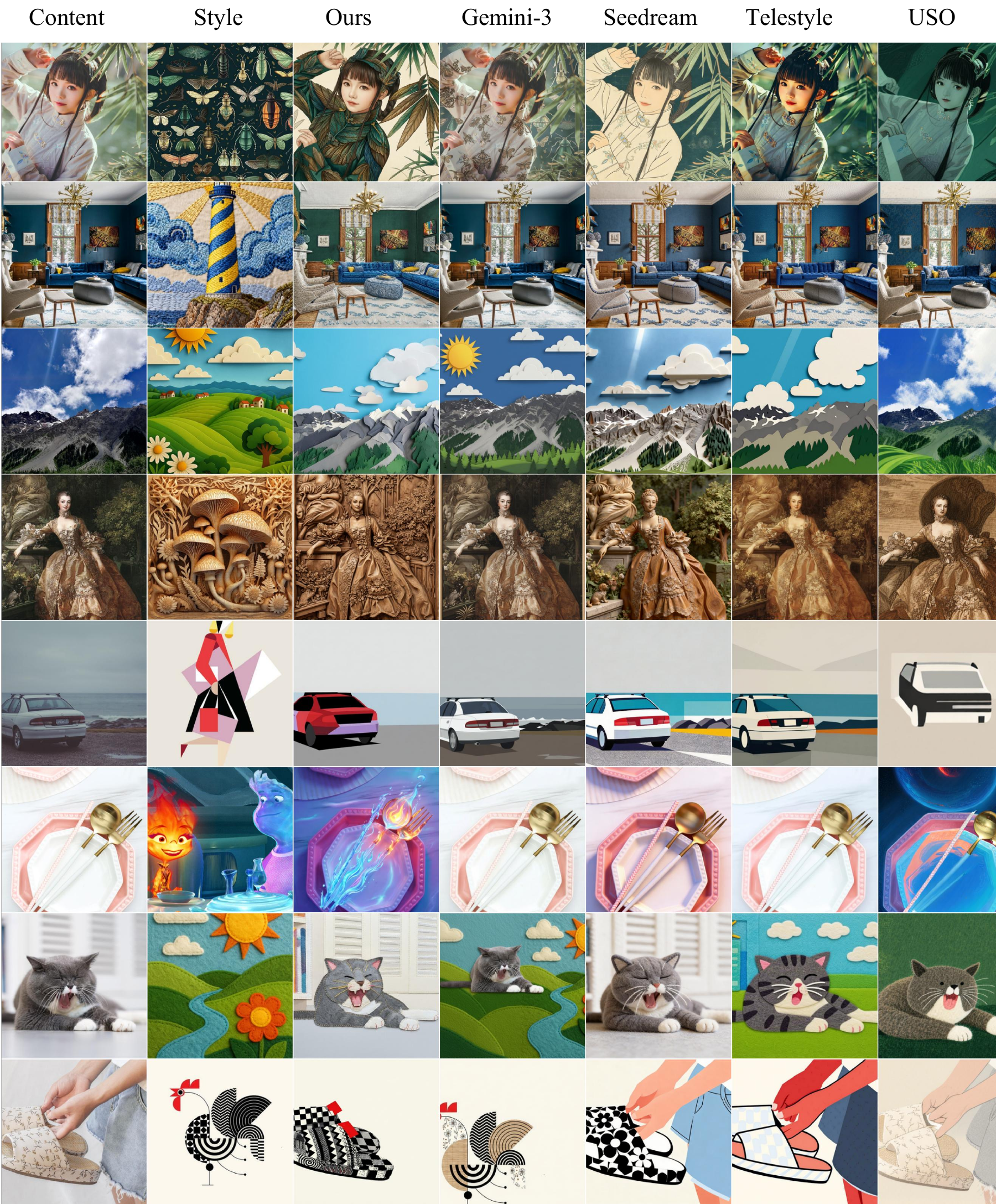}
\caption{\textbf{Qualitative Comparison on Style-reference (\texttt{SRef}) Generation.} Our model achieves faithful stylistic alignment while avoiding the structural artifacts and semantic leakage observed in competing baselines.}
\label{fig:sref_compare_main}
\end{figure*}

\begin{figure*}[tp]
\centering
\includegraphics[width=\linewidth]{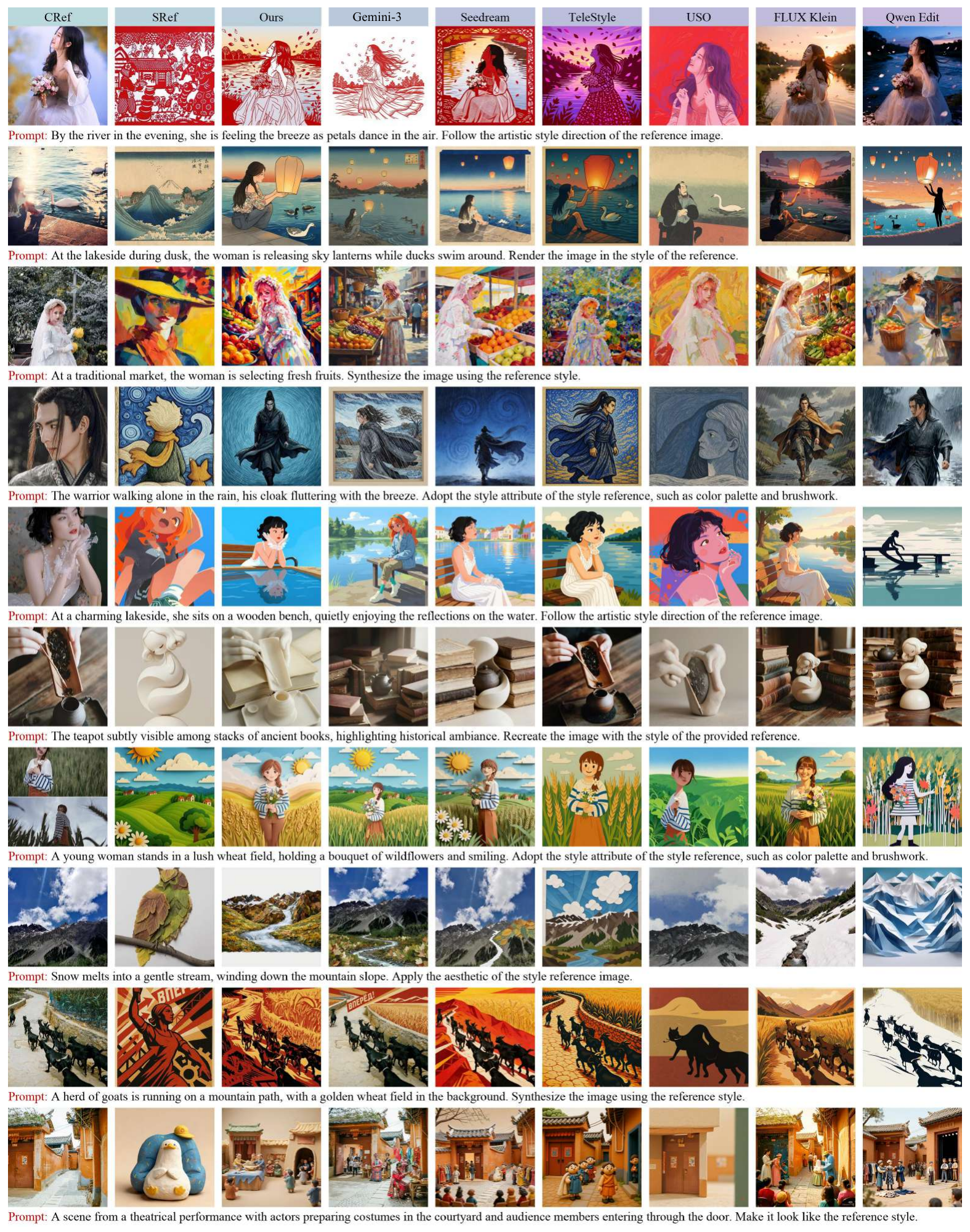}
\vspace{-4ex}
\caption{\small\textbf{Qualitative Comparison on Dual-reference (\texttt{CRef+SRef}) Generation.} Given a content reference (col.~1) and a style reference (col.~2), our method (col.~3) faithfully preserves the layout of the CRef while transferring the artistic attributes of the SRef, exhibiting substantially less semantic leakage than all baselines.}
\label{fig:dualref_compare_main}
\end{figure*}

\paragraph{Qualitative comparison.}
Figure~\ref{fig:sref_compare_main} shows that our method captures the target style (brushwork, texture, palette) with greater precision while preserving the content structure.
Other baselines either introduce structural artifacts or inadvertently copy semantic content from the style reference.
Figure~\ref{fig:dualref_compare_main} extends this comparison to the dual-reference setting, where our method exhibits substantially less semantic leakage than all competing baselines under the same content-style-prompt conditioning.

\subsection{Ablation Studies}\label{sec:ablation}

We validate the three core design choices of FreeStyle: the attention-map constraint, the frequency-aware RoPE modulation, and the training data pipeline.
All ablation variants share the same backbone and training budget, differing only in the component being tested.
Table~\ref{tab:vlm_leakage_ablation} reports VLM-based leakage scores (0--10 scale, lower is better) for the first two components.

\begin{table}[htbp]
\centering
\small
\setlength{\tabcolsep}{6pt}
\renewcommand{\arraystretch}{1.2}
\caption{\textbf{VLM Leakage Score Ablation} (0--10, lower is better). Each row isolates one component by comparing the full model against a variant without it.}
\label{tab:vlm_leakage_ablation}
\begin{tabular}{@{}lcc@{}}
\toprule
\textbf{Component} & \textbf{w/o} & \textbf{w/} \\
\midrule
Enrichment loss (SRef) & 2.674 & 0.522 \\
RoPE modulation (CRef+SRef) & 1.047 & 0.453 \\
\bottomrule
\end{tabular}
\vspace{-10pt}
\end{table}

\begin{figure*}[htbp]
\centering
\includegraphics[width=\linewidth]{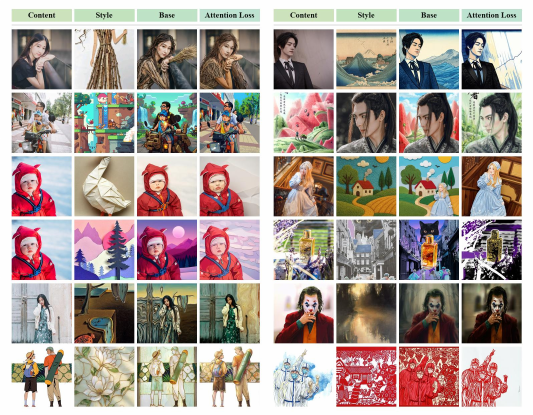}
\caption{\textbf{Ablation on the Attention-map Enrichment Loss.} Without the constraint (left of each pair), the model hallucinates content from the style reference. Enabling it (right) cleanly disentangles style from content.}
\label{fig:loss_ablation}
\end{figure*}

\paragraph{Attention-map constraint.}
Figure~\ref{fig:loss_ablation} illustrates the effect of removing the enrichment loss during Stage~1 training.
Without the constraint, semantic entities from the style reference are frequently hallucinated in the generated image (e.g., objects or architectural elements from the style image appear in the target scene).
Enabling the constraint eliminates this leakage while preserving accurate style transfer.
Quantitatively, the VLM leakage score drops from 2.674 to 0.522, confirming that the attention-level regularization effectively suppresses cross-reference semantic contamination.

\begin{figure*}[htbp]
\centering
\includegraphics[width=0.55\linewidth]{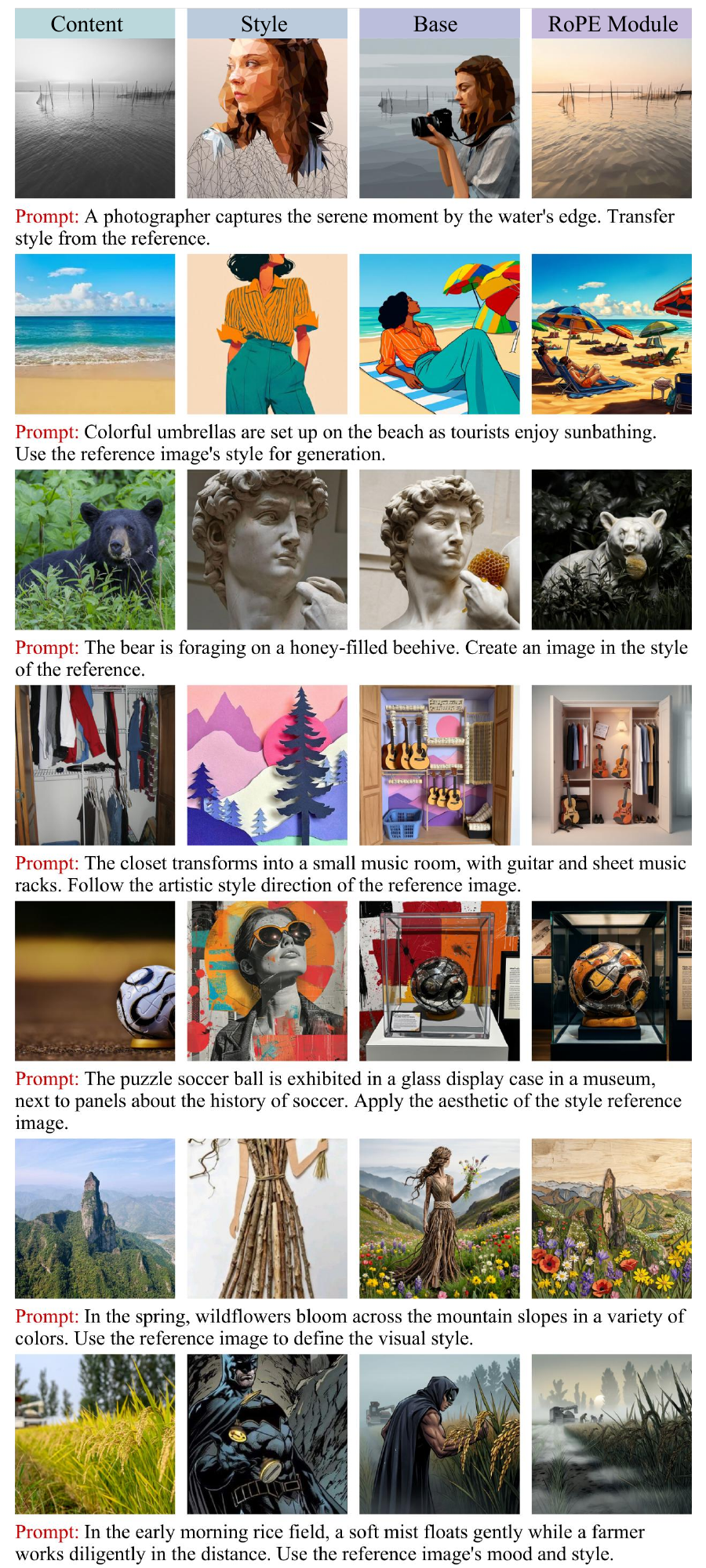}
\caption{\textbf{Ablation on Frequency-aware RoPE Modulation.} Without modulation (left of each pair), positional correspondence causes content leakage from the style reference. Enabling it (right) suppresses leakage while preserving style fidelity.}
\label{fig:rope_ablation}
\end{figure*}

\paragraph{Frequency-aware RoPE modulation.}
Figure~\ref{fig:rope_ablation} compares models trained with and without RoPE modulation under the CRef+SRef setting.
Without modulation, content-specific features from the style image (e.g., object silhouettes or texture patterns) bleed into the output via positional correspondence.
Enabling the modulation suppresses this frequency-domain leakage while maintaining strong style transfer.
The VLM leakage score decreases from 1.047 to 0.453, consistent with the visual comparison.

\begin{figure*}[htbp]
\centering
\includegraphics[width=0.9\linewidth]{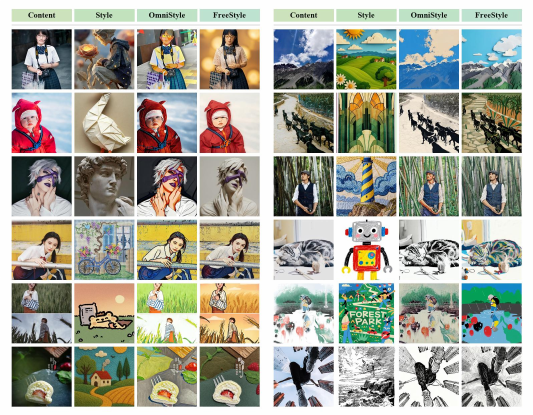}
\caption{\textbf{Dataset Ablation: the Proposed LoRA-based Pipeline vs.\ OmniStyle.} Our model produces more faithful style transfer with richer stylistic detail, whereas the OmniStyle-trained variant shows weaker expressiveness, especially on complex and long-tail styles.}
\label{fig:dataset_ablation}
\end{figure*}

\paragraph{Training data.}
To validate our data pipeline, we compare against a variant trained on OmniStyle data while keeping all other settings identical.
Table~\ref{tab:dataset_ablation} shows clear improvements on style-oriented metrics: ONEIG increases from 0.277 to 0.468, VLM-Style from 3.038 to 7.142, and Ver-S from 0.186 to 0.482.
The qualitative comparison in Figure~\ref{fig:dataset_ablation} reveals an even more pronounced perceptual difference.
Across the displayed examples, the OmniStyle-trained variant often captures only part of the reference appearance, whereas our model more consistently reproduces the overall visual style while preserving the source content.
This qualitative advantage is particularly evident for complex and long-tail styles.

\begin{table}[htbp]
\centering
\small
\setlength{\tabcolsep}{4pt}
\renewcommand{\arraystretch}{1.15}
\caption{\small \textbf{Dataset Ablation on the SRef Benchmark}. Both models use the same architecture and training setup; only the data source differs.}
\label{tab:dataset_ablation}
\begin{tabular}{@{}lcc@{}}
\toprule
\textbf{Metric} & \textbf{OmniStyle} & \textbf{Ours} \\
\midrule
DINOv2 $\uparrow$ & 0.897 & 0.809\\
CAS $\downarrow$ & 0.762 & 1.175\\
ONEIG $\uparrow$ & 0.277 & 0.468\\
CSD $\uparrow$ & 0.577 & 0.639\\
LAION-Aes $\uparrow$ & 5.957 & 6.302\\
V2.5-Aes $\uparrow$ & 5.527 & 5.709\\
VLM-Style $\uparrow$ & 3.038 & 7.142\\
VLM-Content $\uparrow$ & 9.669 & 8.919\\
Ver-S $\uparrow$ & 0.186 & 0.482\\
Ver-C $\uparrow$ & 0.972 & 0.928\\
\bottomrule
\end{tabular}
\end{table}

\section{Discussion and Conclusion}\label{sec:conclusion}
Our results demonstrate that the core challenge of dual-reference generation is not improving style strength or content fidelity in isolation, but stably suppressing cross-reference semantic contamination under multi-condition control.
Experiments and attention analyses confirm that content leakage arises through distinct mechanisms in different settings: attention over-allocation in style-reference generation, and positional correspondence in dual-reference generation.
This motivates our two-stage design where data construction and training constraints are co-optimized for each failure mode.
Several limitations remain.
First, community LoRA quality follows a long-tailed distribution and evolves rapidly, making automated curation an ongoing challenge.
Second, stylistic semantics across different base models still exhibit domain shift, limiting cross-model transferability.
Third, existing evaluation metrics, including the proposed Verification Score, remain insufficient for fine-grained characterization of style-content conflict intensity.
Future work will focus on automated LoRA quality assessment, cross-model style alignment, and more granular leakage metrics.
In summary, we present FreeStyle, a complete framework for style- and content-dual-reference generation comprising a community-LoRA-based data pipeline, a two-stage training strategy with stage-specific disentanglement mechanisms, and a systematic benchmark.
Extensive experiments show that jointly optimizing data, training constraints, and evaluation leads to more robust and balanced dual-reference generation than improving any single component alone.

\section{Data Usage and Ethics Disclaimer}
\label{sec:disclaimer}

This work is conducted strictly for non-commercial academic research purposes. Portions of the data used in this study, including reference images, community LoRA weights, and associated metadata, were collected from publicly accessible websites and open online repositories. These materials were used solely to study and evaluate reference-based image generation, and no part of this work is intended for, or has been deployed in, any commercial product or service.

We do not claim ownership of any third-party content collected during data construction. All copyrights, trademarks, and other intellectual-property rights in the original images, models, and metadata remain with their respective owners. Where applicable, we have made reasonable efforts to respect the terms of use and licenses associated with the source materials, and we use such content only in the limited form necessary for scientific analysis, fair-use academic evaluation, and reproducibility of our reported results. Any names, styles, or trademarks that may appear are the property of their respective holders and are referenced only for identification and research discussion.

The released benchmark, prompts, and code are provided ``as is,'' without any warranty of any kind, express or implied, including but not limited to warranties of merchantability, fitness for a particular purpose, or non-infringement. The authors and their affiliated institutions accept no liability for any direct, indirect, incidental, or consequential damages arising from the use, misuse, or inability to use the data, models, or methods described in this paper. Users who obtain or reproduce any part of this work are solely responsible for ensuring that their own use complies with all applicable laws, regulations, platform terms of service, and third-party rights in their respective jurisdictions.

If any rights holder believes that specific content should not be included, we will promptly remove the corresponding material upon a reasonable request. By using the data, benchmark, or code associated with this paper, users acknowledge and agree to the terms of this disclaimer.

\clearpage
\newpage

\FloatBarrier
{
    \small
    \bibliographystyle{ieeenat_fullname}
    \bibliography{main}

@String(CVPR  = {IEEE Conf. Comput. Vis. Pattern Recog.})

@String(ICCV  = {Int. Conf. Comput. Vis.})

@String(NeurIPS = {Adv. Neural Inform. Process. Syst.})

@String(ICLR  = {Int. Conf. Learn. Represent.})

@String(AAAI  = {AAAI})

@String(CVPR  = {CVPR})

@String(ICCV  = {ICCV})

@String(NeurIPS = {NeurIPS})

@String(ICLR  = {ICLR})

@inproceedings{huang2017adain,
  title={Arbitrary Style Transfer in Real-time with Adaptive Instance Normalization},
  author={Huang, Xun and Belongie, Serge},
  booktitle=ICCV,
  pages={1501--1510},
  year={2017}
}

@inproceedings{li2017wct,
  title={Universal Style Transfer via Feature Transforms},
  author={Li, Yijun and Fang, Chen and Yang, Jimei and Wang, Zhaowen and Lu, Xin and Yang, Ming-Hsuan},
  booktitle=NeurIPS,
  year={2017}
}

@inproceedings{ruiz2023dreambooth,
  title={DreamBooth: Fine Tuning Text-to-Image Diffusion Models for Subject-Driven Generation},
  author={Ruiz, Nataniel and Li, Yuanzhen and Jampani, Varun and Pritch, Yael and Rubinstein, Michael and Aberman, Kfir},
  booktitle=CVPR,
  pages={22500--22510},
  year={2023}
}

@inproceedings{hu2022lora,
  title={LoRA: Low-Rank Adaptation of Large Language Models},
  author={Hu, Edward J and Shen, Yelong and Wallis, Phil and Allen-Zhu, Zeyuan and Li, Yuanzhi and Wang, Shean and Wang, Lu and Chen, Weizhu},
  booktitle=ICLR,
  year={2022}
}

@inproceedings{gal2023textual,
  title={An Image is Worth One Word: Personalizing Text-to-Image Generation using Textual Inversion},
  author={Gal, Rinon and Alaluf, Yuval and Atzmon, Yuval and Patashnik, Or and Bermano, Amit H and Chechik, Gal and Cohen-Or, Daniel},
  booktitle=ICLR,
  year={2023}
}

@article{ye2023ipadapter,
  title={IP-Adapter: Text Compatible Image Prompt Adapter for Text-to-Image Diffusion Models},
  author={Ye, Hu and Zhang, Jun and Liu, Sibo and Han, Xiao and Yang, Wei},
  journal={arXiv preprint arXiv:2308.06721},
  year={2023}
}

@inproceedings{zhang2023controlnet,
  title={Adding Conditional Control to Text-to-Image Diffusion Models},
  author={Zhang, Lvmin and Rao, Anyi and Agrawala, Maneesh},
  booktitle=ICCV,
  pages={3836--3847},
  year={2023}
}

@inproceedings{sohn2023styledrop,
  title={StyleDrop: Text-to-Image Generation in Any Style},
  author={Sohn, Kihyuk and Ruiz, Nataniel and Lee, Kimin and Chin, Daniel Castro and Blok, Irina and Chang, Huiwen and Barber, Jarred and Jiang, Lu and Entis, Glenn and Li, Yuanzhen and others},
  booktitle=NeurIPS,
  year={2023}
}

@misc{laion_aesthetic_predictor,
author       = {{LAION-AI}},
title        = {{LAION-Aesthetics Predictor V1}},
year         = {2022},
howpublished = {\url{https://github.com/LAION-AI/aesthetic-predictor}},
note         = {GitHub repository, accessed June 16, 2026}
}

@article{wang2024instantstyle,
  title={Instantstyle: Free lunch towards style-preserving in text-to-image generation},
  author={Wang, Haofan and Spinelli, Matteo and Wang, Qixun and Bai, Xu and Qin, Zekui and Chen, Anthony},
  journal={arXiv preprint arXiv:2404.02733},
  year={2024}
}

@inproceedings{qi2024deadiff,
  title={Deadiff: An efficient stylization diffusion model with disentangled representations},
  author={Qi, Tianhao and Fang, Shancheng and Wu, Yanze and Xie, Hongtao and Liu, Jiawei and Chen, Lang and He, Qian and Zhang, Yongdong},
  booktitle={Proceedings of the IEEE/CVF conference on computer vision and pattern recognition},
  pages={8693--8702},
  year={2024}
}

@article{xing2024csgo,
  title={CSGO: Content-Style Composition in Text-to-Image Generation},
  author={Xing, Peng and Wang, Haofan and Sun, Yanpeng and Wang, Qixun and Bai, Xu and Ai, Hao and Huang, Renyuan and Li, Zechao},
  journal={arXiv preprint arXiv:2408.16766},
  year={2024}
}

@article{feng2026cleanstyle,
  title={CleanStyle: Plug-and-Play Style Conditioning Purification for Text-to-Image Stylization},
  author={Feng, Xiaoman and Lei, Mingkun and Wang, Yang and Fu, Dingwen and Zhang, Chi},
  journal={arXiv preprint arXiv:2602.20721},
  year={2026}
}

@article{zong2024easyref,
  title={EasyRef: Omni-Generalized Group Image Reference for Diffusion Models via Multimodal LLM},
  author={Zong, Zhuofan and Jiang, Dongzhi and Ma, Bingqi and Song, Guanglu and Shao, Hao and Shen, Dazhong and Liu, Yu and Li, Hongsheng},
  journal={arXiv preprint arXiv:2412.09618},
  year={2024}
}

@article{wang2025omnistyle,
  title={OmniStyle: Filtering High Quality Style Transfer Data at Scale},
  author={Wang, Ye and Liu, Ruiqi and Lin, Jiang and Liu, Fei and Yi, Zili and Wang, Yilin and Ma, Rui},
  journal={arXiv preprint arXiv:2505.14028},
  year={2025}
}

@article{zhang2026telestyle,
  title={TeleStyle: Content-Preserving Style Transfer in Images and Videos},
  author={Zhang, Shiwen and Yang, Xiaoyan and Zi, Bojia and Huang, Haibin and Zhang, Chi and Li, Xuelong},
  journal={arXiv preprint arXiv:2601.20175},
  year={2026}
}

@article{wu2025uso,
  title={USO: Unified Style and Subject-Driven Generation via Disentangled and Reward Learning},
  author={Wu, Shaojin and Huang, Mengqi and Cheng, Yufeng and Wu, Wenxu and Tian, Jiahe and Luo, Yiming and Ding, Fei and He, Qian},
  journal={arXiv preprint arXiv:2508.18966},
  year={2025}
}

@InProceedings{wang2026premier,
    author    = {Wang, Zihao and Wei, Yuxiang and Zhou, Xinpeng and Zhang, Tianyu and Liang, Tao and Bai, Yalong and Zhang, Hongzhi and Zuo, Wangmeng},
    title     = {Premier: Personalized Preference Modulation with Learnable User Embedding in Text-to-Image Generation},
    booktitle = {Proceedings of the IEEE/CVF Conference on Computer Vision and Pattern Recognition (CVPR)},
    month     = {June},
    year      = {2026},
    pages     = {29146-29156}
}

@article{gao2026megastyle,
  title={MegaStyle: Constructing Diverse and Scalable Style Dataset via Consistent Text-to-Image Style Mapping},
  author={Gao, Junyao and Liu, Sibo and Li, Jiaxing and Sun, Yanan and Tu, Yuanpeng and Shen, Fei and Zhang, Weidong and Zhao, Cairong and Zhang, Jun},
  journal={arXiv preprint arXiv:2604.08364},
  year={2026}
}

@misc{openai2025gptimage15,
  author = {{OpenAI}},
  title = {GPT Image 1.5},
  year = {2025},
  howpublished = {\url{https://platform.openai.com/docs/models/gpt-image-1.5}},
  note = {Official model documentation}
}

@misc{google2026nanobananapro,
  author = {{Google DeepMind}},
  title = {Nano Banana Pro},
  year = {2025},
  howpublished = {\url{https://deepmind.google/models/gemini-image/pro/}},
  note = {Official model page}
}

@article{qwen2025qwenimage,
  title={Qwen-Image Technical Report},
  author={{Qwen Team}},
  journal={arXiv preprint arXiv:2508.02324},
  year={2025}
}

@misc{bfl2026flux2klein,
  author = {{Black Forest Labs}},
  title = {FLUX.2 [klein]: Towards Interactive Visual Intelligence},
  year = {2026},
  howpublished = {\url{https://blackforestlabs.ai}},
  note = {Official model announcement, January 15, 2026}
}

@article{ho2020ddpm,
  title={Denoising diffusion probabilistic models},
  author={Ho, Jonathan and Jain, Ajay and Abbeel, Pieter},
  journal={Advances in neural information processing systems},
  volume={33},
  pages={6840--6851},
  year={2020}
}

@article{song2021ddim,
  title={Denoising diffusion implicit models},
  author={Song, Jiaming and Meng, Chenlin and Ermon, Stefano},
  journal={arXiv preprint arXiv:2010.02502},
  year={2020}
}

@article{dhariwal2021diffusion,
  title={Diffusion models beat gans on image synthesis},
  author={Dhariwal, Prafulla and Nichol, Alexander},
  journal={Advances in neural information processing systems},
  volume={34},
  pages={8780--8794},
  year={2021}
}

@article{ho2022cfg,
  title={Classifier-free diffusion guidance},
  author={Ho, Jonathan and Salimans, Tim},
  journal={arXiv preprint arXiv:2207.12598},
  year={2022}
}

@inproceedings{rombach2022ldm,
  title={High-resolution image synthesis with latent diffusion models},
  author={Rombach, Robin and Blattmann, Andreas and Lorenz, Dominik and Esser, Patrick and Ommer, Bj{\"o}rn},
  booktitle={Proceedings of the IEEE/CVF conference on computer vision and pattern recognition},
  pages={10684--10695},
  year={2022}
}

@inproceedings{podell2024sdxl,
  title={Sdxl: Improving latent diffusion models for high-resolution image synthesis},
  author={Podell, Dustin and English, Zion and Lacey, Kyle and Blattmann, Andreas and Dockhorn, Tim and M{\"u}ller, Jonas and Penna, Joe and Rombach, Robin},
  booktitle={International Conference on Learning Representations},
  volume={2024},
  pages={1862--1874},
  year={2024}
}

@article{ramesh2022dalle2,
  title={Hierarchical text-conditional image generation with clip latents},
  author={Ramesh, Aditya and Dhariwal, Prafulla and Nichol, Alex and Chu, Casey and Chen, Mark},
  journal={arXiv preprint arXiv:2204.06125},
  volume={1},
  number={2},
  pages={3},
  year={2022}
}

@article{saharia2022imagen,
  title={Photorealistic text-to-image diffusion models with deep language understanding},
  author={Saharia, Chitwan and Chan, William and Saxena, Saurabh and Li, Lala and Whang, Jay and Denton, Emily L and Ghasemipour, Kamyar and Gontijo Lopes, Raphael and Karagol Ayan, Burcu and Salimans, Tim and others},
  journal={Advances in neural information processing systems},
  volume={35},
  pages={36479--36494},
  year={2022}
}

@article{nichol2022glide,
  title={Glide: Towards photorealistic image generation and editing with text-guided diffusion models},
  author={Nichol, Alex and Dhariwal, Prafulla and Ramesh, Aditya and Shyam, Pranav and Mishkin, Pamela and McGrew, Bob and Sutskever, Ilya and Chen, Mark},
  journal={arXiv preprint arXiv:2112.10741},
  year={2021}
}

@inproceedings{peebles2023dit,
  title={Scalable diffusion models with transformers},
  author={Peebles, William and Xie, Saining},
  booktitle={Proceedings of the IEEE/CVF international conference on computer vision},
  pages={4195--4205},
  year={2023}
}

@inproceedings{esser2024sd3,
  title={Scaling rectified flow transformers for high-resolution image synthesis},
  author={Esser, Patrick and Kulal, Sumith and Blattmann, Andreas and Entezari, Rahim and M{\"u}ller, Jonas and Saini, Harry and Levi, Yam and Lorenz, Dominik and Sauer, Axel and Boesel, Frederic and others},
  booktitle={Forty-first international conference on machine learning},
  year={2024}
}

@article{lipman2023flowmatching,
  title={Flow matching for generative modeling},
  author={Lipman, Yaron and Chen, Ricky TQ and Ben-Hamu, Heli and Nickel, Maximilian and Le, Matt},
  journal={arXiv preprint arXiv:2210.02747},
  year={2022}
}

@inproceedings{liu2023rectifiedflow,
  title={Flow Straight and Fast: Learning to Generate and Transfer Data with Rectified Flow},
  author={Liu, Xingchao and Gong, Chengyue and Liu, Qiang},
  booktitle=ICLR,
  year={2023}
}

@article{vaswani2017attention,
  title={Attention is all you need},
  author={Vaswani, Ashish and Shazeer, Noam and Parmar, Niki and Uszkoreit, Jakob and Jones, Llion and Gomez, Aidan N and Kaiser, {\L}ukasz and Polosukhin, Illia},
  journal={Advances in neural information processing systems},
  volume={30},
  year={2017}
}

@article{dosovitskiy2021vit,
  title={An image is worth 16x16 words: Transformers for image recognition at scale},
  author={Dosovitskiy, Alexey and Beyer, Lucas and Kolesnikov, Alexander and Weissenborn, Dirk and Zhai, Xiaohua and Unterthiner, Thomas and Dehghani, Mostafa and Minderer, Matthias and Heigold, Georg and Gelly, Sylvain and others},
  journal={arXiv preprint arXiv:2010.11929},
  year={2020}
}

@article{su2024roformer,
  title={Roformer: Enhanced transformer with rotary position embedding},
  author={Su, Jianlin and Ahmed, Murtadha and Lu, Yu and Pan, Shengfeng and Bo, Wen and Liu, Yunfeng},
  journal={Neurocomputing},
  volume={568},
  pages={127063},
  year={2024},
  publisher={Elsevier}
}

@inproceedings{gatys2016style,
  title={Image style transfer using convolutional neural networks},
  author={Gatys, Leon A and Ecker, Alexander S and Bethge, Matthias},
  booktitle={Proceedings of the IEEE conference on computer vision and pattern recognition},
  pages={2414--2423},
  year={2016}
}

@inproceedings{johnson2016perceptual,
  title={Perceptual losses for real-time style transfer and super-resolution},
  author={Johnson, Justin and Alahi, Alexandre and Fei-Fei, Li},
  booktitle={European conference on computer vision},
  pages={694--711},
  year={2016},
  organization={Springer}
}

@inproceedings{zhu2017cyclegan,
  title={Unpaired Image-to-Image Translation Using Cycle-Consistent Adversarial Networks},
  author={Zhu, Jun-Yan and Park, Taesung and Isola, Phillip and Efros, Alexei A},
  booktitle=ICCV,
  pages={2223--2232},
  year={2017}
}

@inproceedings{karras2019stylegan,
  title={A style-based generator architecture for generative adversarial networks},
  author={Karras, Tero and Laine, Samuli and Aila, Timo},
  booktitle={Proceedings of the IEEE/CVF conference on computer vision and pattern recognition},
  pages={4401--4410},
  year={2019}
}

@inproceedings{radford2021clip,
  title={Learning transferable visual models from natural language supervision},
  author={Radford, Alec and Kim, Jong Wook and Hallacy, Chris and Ramesh, Aditya and Goh, Gabriel and Agarwal, Sandhini and Sastry, Girish and Askell, Amanda and Mishkin, Pamela and Clark, Jack and others},
  booktitle={International conference on machine learning},
  pages={8748--8763},
  year={2021},
  organization={PmLR}
}

@article{oquab2023dinov2,
  title={Dinov2: Learning robust visual features without supervision},
  author={Oquab, Maxime and Darcet, Timoth{\'e}e and Moutakanni, Th{\'e}o and Vo, Huy and Szafraniec, Marc and Khalidov, Vasil and Fernandez, Pierre and Haziza, Daniel and Massa, Francisco and El-Nouby, Alaaeldin and others},
  journal={arXiv preprint arXiv:2304.07193},
  year={2023}
}

@article{somepalli2024csd,
  title={Measuring style similarity in diffusion models},
  author={Somepalli, Gowthami and Gupta, Anubhav and Gupta, Kamal and Palta, Shramay and Goldblum, Micah and Geiping, Jonas and Shrivastava, Abhinav and Goldstein, Tom},
  journal={arXiv preprint arXiv:2404.01292},
  year={2024}
}

@inproceedings{kumari2023customdiffusion,
  title={Multi-concept customization of text-to-image diffusion},
  author={Kumari, Nupur and Zhang, Bingliang and Zhang, Richard and Shechtman, Eli and Zhu, Jun-Yan},
  booktitle={Proceedings of the IEEE/CVF conference on computer vision and pattern recognition},
  pages={1931--1941},
  year={2023}
}

@inproceedings{brooks2023instructpix2pix,
  title={Instructpix2pix: Learning to follow image editing instructions},
  author={Brooks, Tim and Holynski, Aleksander and Efros, Alexei A},
  booktitle={Proceedings of the IEEE/CVF conference on computer vision and pattern recognition},
  pages={18392--18402},
  year={2023}
}

@inproceedings{
hertz2023prompt2prompt,
title={Prompt-to-Prompt Image Editing with Cross-Attention Control},
author={Amir Hertz and Ron Mokady and Jay Tenenbaum and Kfir Aberman and Yael Pritch and Daniel Cohen-Or},
booktitle={The Eleventh International Conference on Learning Representations },
year={2023},
url={https://openreview.net/forum?id=_CDixzkzeyb}
}

@inproceedings{
meng2022sdedit,
title={{SDE}dit: Guided Image Synthesis and Editing with Stochastic Differential Equations},
author={Chenlin Meng and Yutong He and Yang Song and Jiaming Song and Jiajun Wu and Jun-Yan Zhu and Stefano Ermon},
booktitle={International Conference on Learning Representations},
year={2022},
url={https://openreview.net/forum?id=aBsCjcPu_tE}
}

@inproceedings{chung2024styleid,
  title={Style injection in diffusion: A training-free approach for adapting large-scale diffusion models for style transfer},
  author={Chung, Jiwoo and Hyun, Sangeek and Heo, Jae-Pil},
  booktitle={Proceedings of the IEEE/CVF conference on computer vision and pattern recognition},
  pages={8795--8805},
  year={2024}
}

@article{qwen3vl,
  title={Qwen3-vl technical report},
  author={Bai, Shuai and Cai, Yuxuan and Chen, Ruizhe and Chen, Keqin and Chen, Xionghui and Cheng, Zesen and Deng, Lianghao and Ding, Wei and Gao, Chang and Ge, Chunjiang and others},
  journal={arXiv preprint arXiv:2511.21631},
  year={2025}
}

@inproceedings{mou2024t2iadapter,
  title={T2i-adapter: Learning adapters to dig out more controllable ability for text-to-image diffusion models},
  author={Mou, Chong and Wang, Xintao and Xie, Liangbin and Wu, Yanze and Zhang, Jian and Qi, Zhongang and Shan, Ying},
  booktitle={Proceedings of the AAAI conference on artificial intelligence},
  volume={38},
  number={5},
  pages={4296--4304},
  year={2024}
}

@article{mikaeili2026untwistingrope,
  title={Untwisting RoPE: Frequency Control for Shared Attention in DiTs},
  author={Mikaeili, Aryan and Patashnik, Or and Tagliasacchi, Andrea and Cohen-Or, Daniel and Mahdavi-Amiri, Ali},
  journal={arXiv preprint arXiv:2602.05013},
  year={2026}
}

@article{ulyanov2016texture,
  title={Texture networks: Feed-forward synthesis of textures and stylized images},
  author={Ulyanov, Dmitry and Lebedev, Vadim and Vedaldi, Andrea and Lempitsky, Victor},
  journal={arXiv preprint arXiv:1603.03417},
  year={2016}
}

@inproceedings{
dumoulin2017learned,
title={A Learned Representation For Artistic Style},
author={Vincent Dumoulin and Jonathon Shlens and Manjunath Kudlur},
booktitle={International Conference on Learning Representations},
year={2017},
url={https://openreview.net/forum?id=BJO-BuT1g}
}

@article{chenschmidt2016styleswap,
  title={Fast patch-based style transfer of arbitrary style},
  author={Chen, Tian Qi and Schmidt, Mark},
  journal={arXiv preprint arXiv:1612.04337},
  year={2016}
}

@inproceedings{luan2017deepphoto,
  title={Deep photo style transfer},
  author={Luan, Fujun and Paris, Sylvain and Shechtman, Eli and Bala, Kavita},
  booktitle={Proceedings of the IEEE conference on computer vision and pattern recognition},
  pages={4990--4998},
  year={2017}
}

@inproceedings{sheng2018avatarnet,
  title={Avatar-net: Multi-scale zero-shot style transfer by feature decoration},
  author={Sheng, Lu and Lin, Ziyi and Shao, Jing and Wang, Xiaogang},
  booktitle={Proceedings of the IEEE conference on computer vision and pattern recognition},
  pages={8242--8250},
  year={2018}
}

@inproceedings{park2019sanet,
  title={Arbitrary style transfer with style-attentional networks},
  author={Park, Dae Young and Lee, Kwang Hee},
  booktitle={proceedings of the IEEE/CVF conference on computer vision and pattern recognition},
  pages={5880--5888},
  year={2019}
}

@inproceedings{an2021artflow,
  title={Artflow: Unbiased image style transfer via reversible neural flows},
  author={An, Jie and Huang, Siyu and Song, Yibing and Dou, Dejing and Liu, Wei and Luo, Jiebo},
  booktitle={Proceedings of the IEEE/CVF conference on computer vision and pattern recognition},
  pages={862--871},
  year={2021}
}

@inproceedings{liu2021adaattn,
  title={Adaattn: Revisit attention mechanism in arbitrary neural style transfer},
  author={Liu, Songhua and Lin, Tianwei and He, Dongliang and Li, Fu and Wang, Meiling and Li, Xin and Sun, Zhengxing and Li, Qian and Ding, Errui},
  booktitle={Proceedings of the IEEE/CVF international conference on computer vision},
  pages={6649--6658},
  year={2021}
}

@inproceedings{deng2022stytr2,
  title={Stytr2: Image style transfer with transformers},
  author={Deng, Yingying and Tang, Fan and Dong, Weiming and Ma, Chongyang and Pan, Xingjia and Wang, Lei and Xu, Changsheng},
  booktitle={Proceedings of the IEEE/CVF conference on computer vision and pattern recognition},
  pages={11326--11336},
  year={2022}
}

@inproceedings{zhang2022efdm,
  title={Exact feature distribution matching for arbitrary style transfer and domain generalization},
  author={Zhang, Yabin and Li, Minghan and Li, Ruihuang and Jia, Kui and Zhang, Lei},
  booktitle={Proceedings of the IEEE/CVF conference on computer vision and pattern recognition},
  pages={8035--8045},
  year={2022}
}

@inproceedings{wu2022ccpl,
  title={Ccpl: Contrastive coherence preserving loss for versatile style transfer},
  author={Wu, Zijie and Zhu, Zhen and Du, Junping and Bai, Xiang},
  booktitle={European conference on computer vision},
  pages={189--206},
  year={2022},
  organization={Springer}
}

@article{chen2021ieconst,
  title={Artistic style transfer with internal-external learning and contrastive learning},
  author={Chen, Haibo and Wang, Zhizhong and Zhang, Huiming and Zuo, Zhiwen and Li, Ailin and Xing, Wei and Lu, Dongming and others},
  journal={Advances in Neural Information Processing Systems},
  volume={34},
  pages={26561--26573},
  year={2021}
}

@inproceedings{yoo2019wct2,
  title={Photorealistic style transfer via wavelet transforms},
  author={Yoo, Jaejun and Uh, Youngjung and Chun, Sanghyuk and Kang, Byeongkyu and Ha, Jung-Woo},
  booktitle={Proceedings of the IEEE/CVF international conference on computer vision},
  pages={9036--9045},
  year={2019}
}

@inproceedings{li2019linear,
  title={Learning linear transformations for fast image and video style transfer},
  author={Li, Xueting and Liu, Sifei and Kautz, Jan and Yang, Ming-Hsuan},
  booktitle={Proceedings of the IEEE/CVF conference on computer vision and pattern recognition},
  pages={3809--3817},
  year={2019}
}

@inproceedings{kolkin2019strotss,
  title={Style transfer by relaxed optimal transport and self-similarity},
  author={Kolkin, Nicholas and Salavon, Jason and Shakhnarovich, Gregory},
  booktitle={Proceedings of the IEEE/CVF conference on computer vision and pattern recognition},
  pages={10051--10060},
  year={2019}
}

@article{seedream2025seedream,
  title={Seedream 4.0: Toward next-generation multimodal image generation},
  author={Seedream, Team and Chen, Yunpeng and Gao, Yu and Gong, Lixue and Guo, Meng and Guo, Qiushan and Guo, Zhiyao and Hou, Xiaoxia and Huang, Weilin and Huang, Yixuan and others},
  journal={arXiv preprint arXiv:2509.20427},
  year={2025}
}

@misc{verb2024aesthetic,
  author       = {Verb},
  title        = {{Aesthetic Predictor V2.5}: A {SigLIP}-Based Aesthetic Score Predictor},
  year         = {2024},
  howpublished = {\url{https://github.com/discus0434/aesthetic-predictor-v2-5}},
  note         = {Version 2024.12.18.1, GitHub repository}
}

@inproceedings{zhang2023inst,
  title={Inversion-based style transfer with diffusion models},
  author={Zhang, Yuxin and Huang, Nisha and Tang, Fan and Huang, Haibin and Ma, Chongyang and Dong, Weiming and Xu, Changsheng},
  booktitle={Proceedings of the IEEE/CVF conference on computer vision and pattern recognition},
  pages={10146--10156},
  year={2023}
}

@article{chang2026oneig,
  title={Oneig-bench: Omni-dimensional nuanced evaluation for image generation},
  author={Chang, Jingjing and Fang, Yixiao and Xing, Peng and Wu, Shuhan and Cheng, Wei and Wang, Rui and Zeng, Xianfang and Yu, Gang and Chen, Hai-Bao},
  journal={Advances in Neural Information Processing Systems},
  volume={38},
  year={2026}
}

@inproceedings{hertz2024stylealigned,
  title={Style aligned image generation via shared attention},
  author={Hertz, Amir and Voynov, Andrey and Fruchter, Shlomi and Cohen-Or, Daniel},
  booktitle={Proceedings of the IEEE/CVF Conference on Computer Vision and Pattern Recognition},
  pages={4775--4785},
  year={2024}
}

@article{jeong2024visualstyle,
  title={Visual style prompting with swapping self-attention},
  author={Jeong, Jaeseok and Kim, Junho and Choi, Yunjey and Lee, Gayoung and Uh, Youngjung},
  journal={arXiv preprint arXiv:2402.12974},
  year={2024}
}

@inproceedings{shah2024ziplora,
  title={Ziplora: Any subject in any style by effectively merging loras},
  author={Shah, Viraj and Ruiz, Nataniel and Cole, Forrester and Lu, Erika and Lazebnik, Svetlana and Li, Yuanzhen and Jampani, Varun},
  booktitle={European Conference on Computer Vision},
  pages={422--438},
  year={2024},
  organization={Springer}
}

@inproceedings{frenkel2024blora,
  title={Implicit style-content separation using b-lora},
  author={Frenkel, Yarden and Vinker, Yael and Shamir, Ariel and Cohen-Or, Daniel},
  booktitle={European Conference on Computer Vision},
  pages={181--198},
  year={2024},
  organization={Springer}
}

@article{goodfellow2014gan,
  title={Generative adversarial nets},
  author={Goodfellow, Ian J and Pouget-Abadie, Jean and Mirza, Mehdi and Xu, Bing and Warde-Farley, David and Ozair, Sherjil and Courville, Aaron and Bengio, Yoshua},
  journal={Advances in neural information processing systems},
  volume={27},
  year={2014}
}

@inproceedings{karras2020stylegan2,
  title={Analyzing and improving the image quality of stylegan},
  author={Karras, Tero and Laine, Samuli and Aittala, Miika and Hellsten, Janne and Lehtinen, Jaakko and Aila, Timo},
  booktitle={Proceedings of the IEEE/CVF conference on computer vision and pattern recognition},
  pages={8110--8119},
  year={2020}
}

@misc{civitai,
author = {{Civitai}},
title = {Civitai: A Generative AI Model-Sharing Platform},
howpublished = {\url{https://civitai.com/}},
note = {Accessed: June 16, 2026}
}

@misc{tensorart,
author = {{Tensor.Art}},
title = {Tensor.Art: An AI Model-Sharing and Generation Platform},
howpublished = {\url{https://tensor.art/}},
note = {Accessed: June 16, 2026}
}

@misc{liblibai,
author = {{LiblibAI}},
title = {LiblibAI: An AI Creation and Model-Sharing Platform},
howpublished = {\url{https://www.liblib.art/}},
note = {Accessed: June 16, 2026}
}

@article{bai2025qwen25vl,
title   = {Qwen2.5-VL Technical Report},
author  = {Bai, Shuai and Chen, Keqin and Liu, Xuejing and Wang, Jialin and Ge, Wenbin and Song, Sibo and Dang, Kai and Wang, Peng and Wang, Shijie and Tang, Jun and Zhong, Humen and Zhu, Yuanzhi and Yang, Mingkun and Li, Zhaohai and Wan, Jianqiang and Wang, Pengfei and Ding, Wei and Fu, Zheren and Xu, Yiheng and Ye, Jiabo and Zhang, Xi and Xie, Tianbao and Cheng, Zesen and Zhang, Hang and Yang, Zhibo and Xu, Haiyang and Lin, Junyang},
journal = {arXiv preprint arXiv:2502.13923},
year    = {2025}
}

@article{park2024illustrious,
  title={Illustrious: an open advanced illustration model},
  author={Park, Sang Hyun and Koh, Jun Young and Lee, Junha and Song, Joy and Kim, Dongha and Moon, Hoyeon and Lee, Hyunju and Song, Min},
  journal={arXiv preprint arXiv:2409.19946},
  year={2024}
}

@misc{comfyanonymous2023comfyui,
  author={{comfyanonymous}},
  title={{ComfyUI}: A Powerful and Modular Stable Diffusion GUI and Backend},
  year={2023},
  howpublished={\url{https://github.com/comfyanonymous/ComfyUI}},
  note={GitHub repository}
}

@article{kuznetsova2020openimage,
  title={The open images dataset v4: Unified image classification, object detection, and visual relationship detection at scale},
  author={Kuznetsova, Alina and Rom, Hassan and Alldrin, Neil and Uijlings, Jasper and Krasin, Ivan and Pont-Tuset, Jordi and Kamali, Shahab and Popov, Stefan and Malloci, Matteo and Kolesnikov, Alexander and others},
  journal={International journal of computer vision},
  volume={128},
  number={7},
  pages={1956--1981},
  year={2020},
  publisher={Springer}
}

@article{openai2023gpt4,
  title={Gpt-4 technical report},
  author={Achiam, Josh and Adler, Steven and Agarwal, Sandhini and Ahmad, Lama and Akkaya, Ilge and Aleman, Florencia Leoni and Almeida, Diogo and Altenschmidt, Janko and Altman, Sam and Anadkat, Shyamal and others},
  journal={arXiv preprint arXiv:2303.08774},
  year={2023}
}

@misc{flux2024,
    author={Black Forest Labs},
    title={FLUX},
    year={2024},
    howpublished={\url{https://github.com/black-forest-labs/flux}},
}

@article{fu2025imontage,
  title={iMontage: Unified, Versatile, Highly Dynamic Many-to-many Image Generation},
  author={Fu, Zhoujie and Zeng, Xianfang and Lan, Jinghong and Liao, Xinyao and Chen, Cheng and Chen, Junyi and Wei, Jiacheng and Cheng, Wei and Liu, Shiyu and Chen, Yunuo and others},
  journal={arXiv preprint arXiv:2511.20635},
  year={2025}
}
}

\clearpage
\newpage
\appendix
\section{Appendix}

\subsection{Model Architecture without Extra Image Encoders}

Our style reference integration does not rely on an additional external image encoder. This design choice is motivated by the observation that existing image encoders possess limited capabilities in clustering and classifying artistic styles, and relying on them for style feature extraction might ultimately impair the model's performance. Therefore, during training, we disentangle the model's attention layers and apply our constraints directly on the first block. The overall architecture of our model is summarized in the middle panel of Figure~\ref{fig:attention_enrichment_combined}.

\begin{figure*}[t]
\centering
\includegraphics[width=0.8\linewidth]{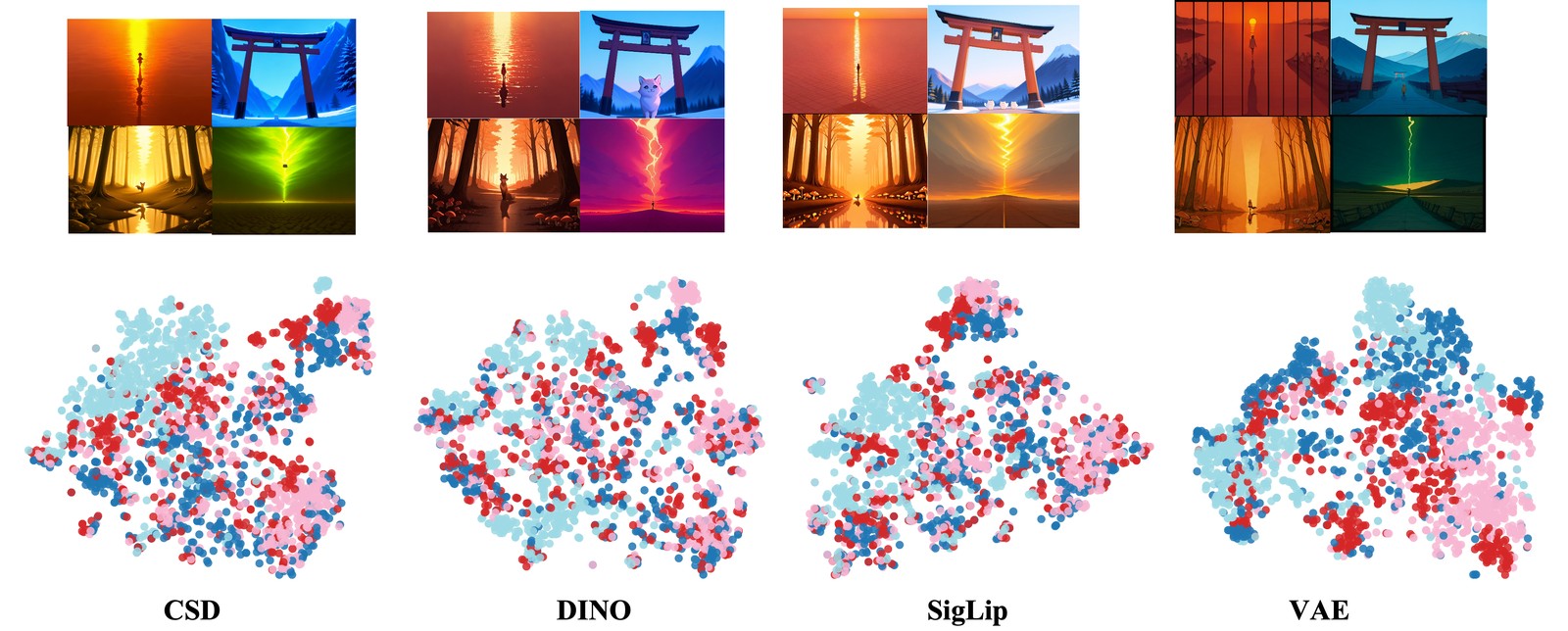}
\caption{\textbf{t-SNE Visualization of Style Clustering in Different Encoder Feature Spaces.} Test images are synthesized from 4 distinct style LoRAs using the same 150 prompts, with each LoRA serving as a style cluster center. Compared with CLIP and DINOv2, VAE latent features yield the clearest separation of style clusters, motivating our encoder-free design that directly guides generation with VAE latents.}
\label{fig:tsne_vit}
\vspace{-5mm}
\end{figure*}

To further illustrate why we do not employ other external image encoders, we conducted a clustering experiment on stylistic features. As shown in Figure~\ref{fig:tsne_vit}, we selected 4 different style LoRAs and combined their respective trigger words with the same set of 150 base prompts to generate test images via text-to-image synthesis. Since style LoRAs can be viewed as clustering centers for different stylistic concepts, we extracted the features of these generated images using several common image encoders (e.g., CLIP, DINOv2) as well as the VAE, and visualized them using t-SNE dimensionality reduction. The observation reveals that compared to other image encoders, the features extracted by the VAE can relatively clearly distinguish the clusters of different styles. This further validates the rationale behind our architectural choice to directly utilize the latent features extracted by the VAE for explicit guidance.

\subsection{Why Constrain the First Block}

\begin{figure}[t]
\centering
\includegraphics[width=\linewidth]{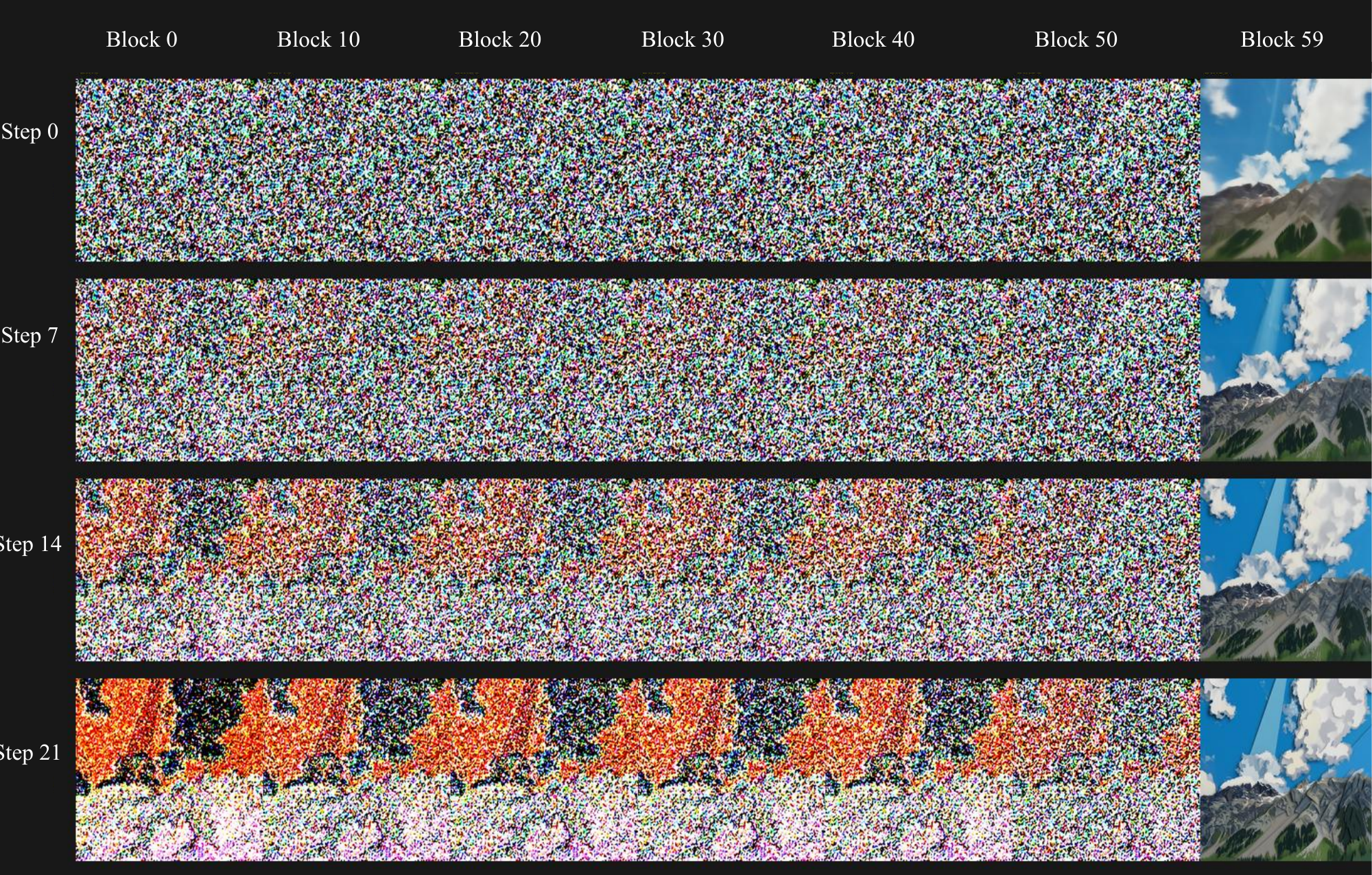}
\caption{\textbf{Semantic Analysis of Intermediate Features across Transformer Blocks and Denoising Steps.} Early blocks (block~0) primarily encode global semantics and layout, whereas later blocks focus on fine-grained details. As semantic leakage at block~0 cannot be corrected by subsequent layers, we impose our attention constraint at this stage.}
\label{fig:noise_frequency}
\end{figure}

To further motivate our design choice of applying the attention constraint on the first block, Figure~\ref{fig:noise_frequency} visualizes the intermediate feature representations across different denoising steps and block indices. It can be clearly observed that early blocks (especially the first block) determine the overall spatial layout and semantic composition of the generated image at the very beginning of the denoising process. As the block index increases, features progressively refine texture and local details, but the global structure has already been established in the early blocks. This observation indicates that if content leakage occurs at the first-block level, it will directly corrupt the semantic layout of the entire generated image, and subsequent blocks can hardly correct it. Therefore, applying the attention constraint on the first block is the most effective intervention point for suppressing cross-reference semantic leakage.

\subsection{Trigger-Word Statistics}
This figure reports phrase-level word cloud of the LoRA trigger words for style and content, providing additional evidence for the vocabulary diversity used in data construction.
\begin{figure*}[b]
\centering
\begin{minipage}[t]{0.49\linewidth}
\centering
\includegraphics[width=\linewidth,height=0.24\textheight,keepaspectratio]{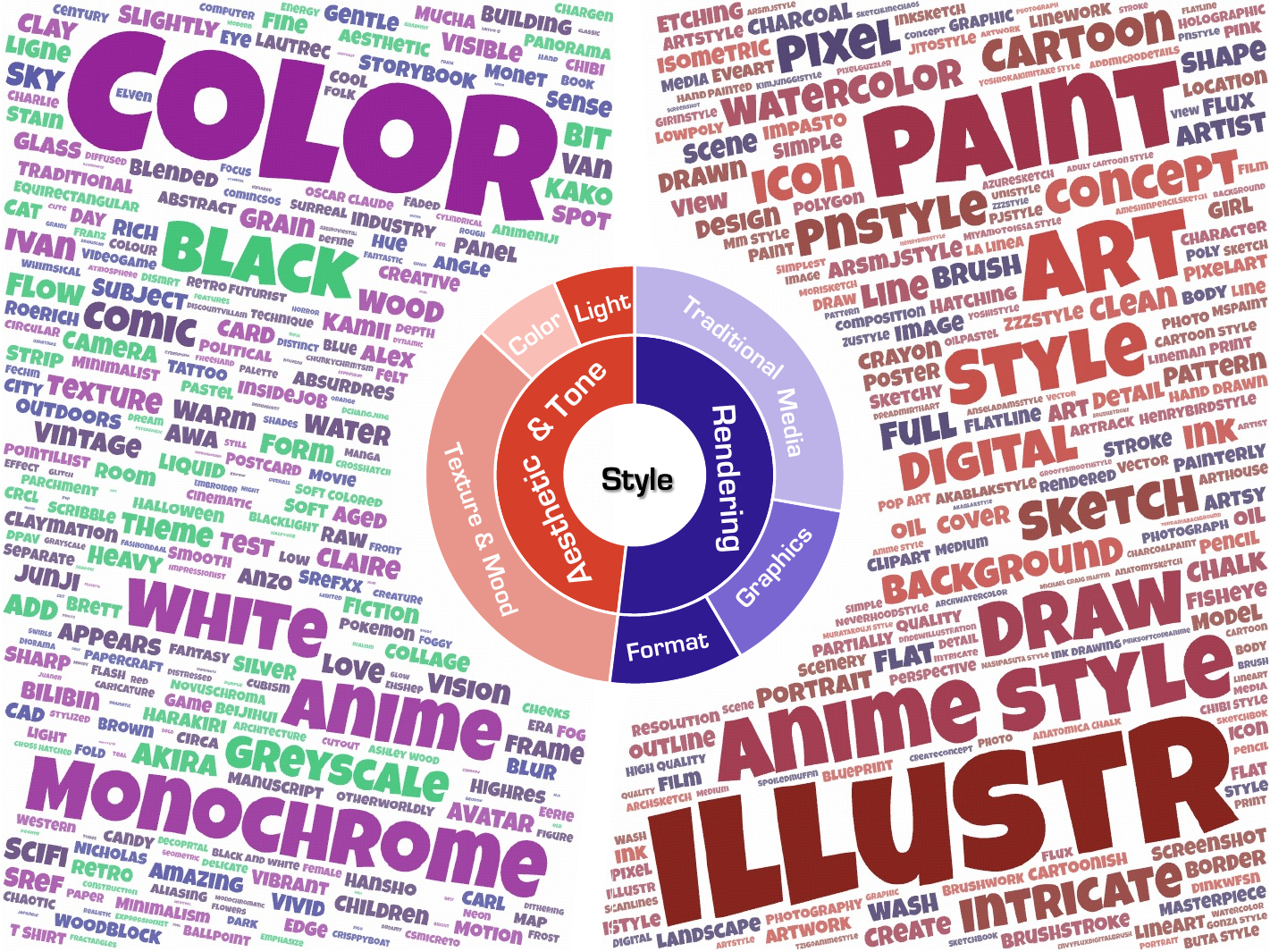}
\end{minipage}\hfill
\begin{minipage}[t]{0.49\linewidth}
\centering
\includegraphics[width=\linewidth,height=0.24\textheight,keepaspectratio]{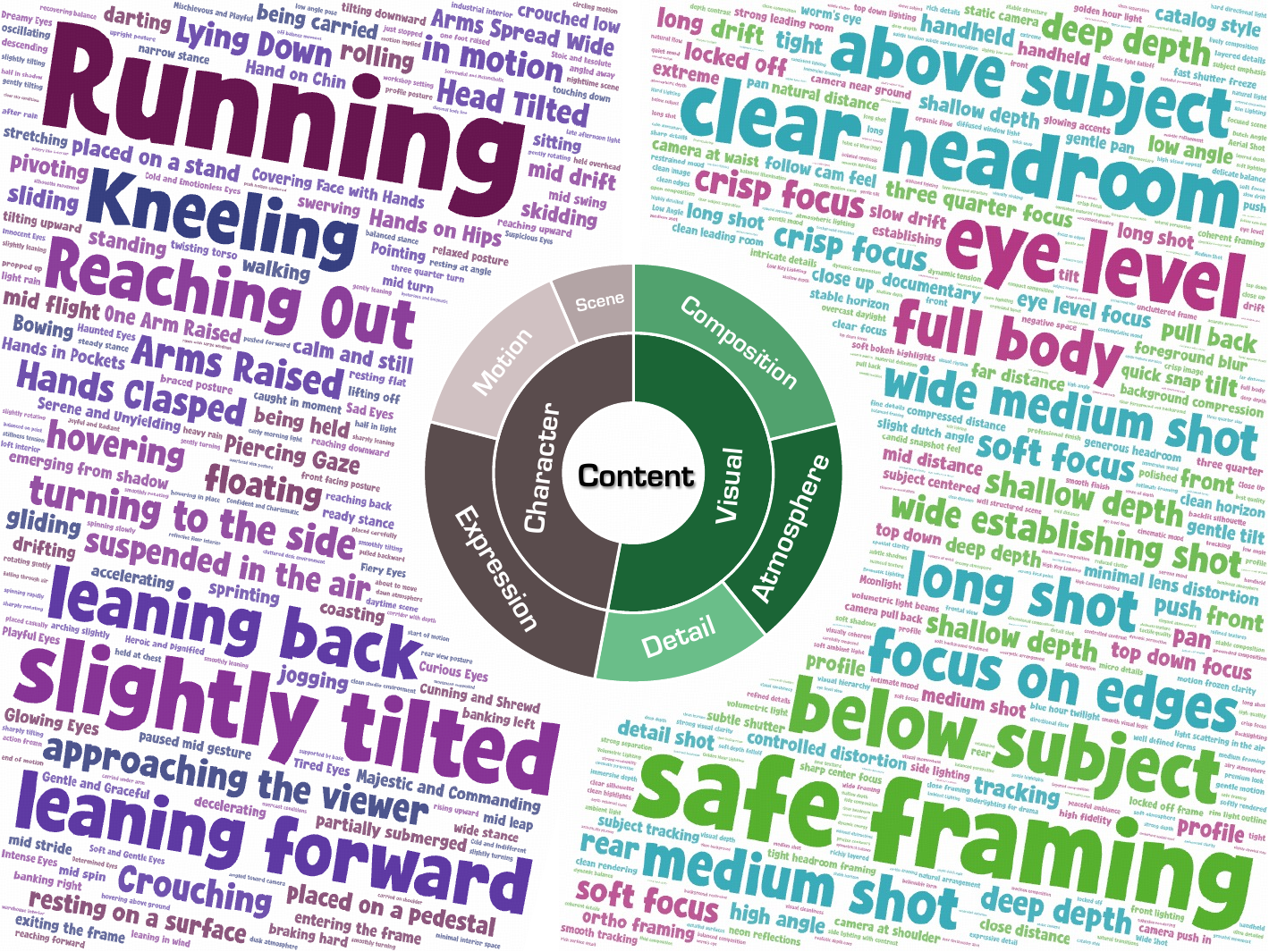}
\end{minipage}
\caption{\textbf{Trigger-word Word Clouds in the Collected LoRA Metadata.} The style (left) and content (right) word clouds together evidence the vocabulary diversity underlying our prompt-pool construction.}
\label{fig:trigger_word_distribution}
\end{figure*}

\subsection{CRef and SRef Dataset Composition}

To elaborate on the scale of our constructed SRef and CRef dataset, Table~\ref{tab:dataset_composition} reports the detailed data composition across the three base models (Qwen, FLUX, and Illustrious). The total number of triplets is calculated as the product of the retained dual-reference LoRA combinations and the average number of generated images per combination, per style LoRA, and per content LoRA. For instance, in the Qwen setting, we retain 608 valid LoRA combinations (with an average of 8.85 images per combination). Combined with 53 style LoRAs (averaging 8.00 images) and 19 content LoRAs (averaging 13.12 images).This multiplicative compositional mechanism guarantees immense diversity at both the content and stylistic levels.

\begin{table}[t]
\centering
\caption{\textbf{Detailed Statistics of the Curated CRef and SRef Dataset Composition across Different Base Models}. The total number of triplets is derived by multiplying the number of dual LoRA combinations, style LoRAs, and content LoRAs by their respective average number of generated images.The “Total Triplets” reported in the table are calculated based on the Cartesian product. However, the actual number of samples released is obtained by sufficiently sampling the style reference images rather than exhaustively enumerating all possible combinations.}
\vspace{0.1cm}
\scriptsize
\setlength{\tabcolsep}{2.5pt}
\renewcommand{\arraystretch}{1.05}
\resizebox{\columnwidth}{!}{
\begin{tabular}{@{}l|cc|cc|cc|c@{}}
\toprule
\multirow{2}{*}{\textbf{Base Model}} & \multicolumn{2}{c|}{\shortstack{\textbf{Dual LoRA}\\\textbf{Combinations}}} & \multicolumn{2}{c|}{\shortstack{\textbf{Style}\\\textbf{LoRAs}}} & \multicolumn{2}{c|}{\shortstack{\textbf{Content}\\\textbf{LoRAs}}} & \multirow{2}{*}{\shortstack{\textbf{Total}\\\textbf{Triplets}}} \\
\cmidrule(lr){2-3} \cmidrule(lr){4-5} \cmidrule(lr){6-7}
 & Count & Avg. Images & Count & Avg. Images & Count & Avg. Images & \\
\midrule
Qwen & 608 & 8.85 & 53 & 8.00 & 19 & 13.12 & $\sim$935K \\
FLUX & 43,750 & 3.50 & 1,460 & 9.11 & 91 & 9.16 & $\sim$38.80M \\
Illustrious & 24,127 & 3.15 & 191 & 9.36 & 799 & 15.20 & $\sim$68.39M \\
\bottomrule
\end{tabular}
}
\label{tab:dataset_composition}
\end{table}

\subsection{Additional Visual Comparisons for SRef}

\begin{figure*}[p]
\centering
\includegraphics[width=\textwidth,height=0.93\textheight,keepaspectratio]{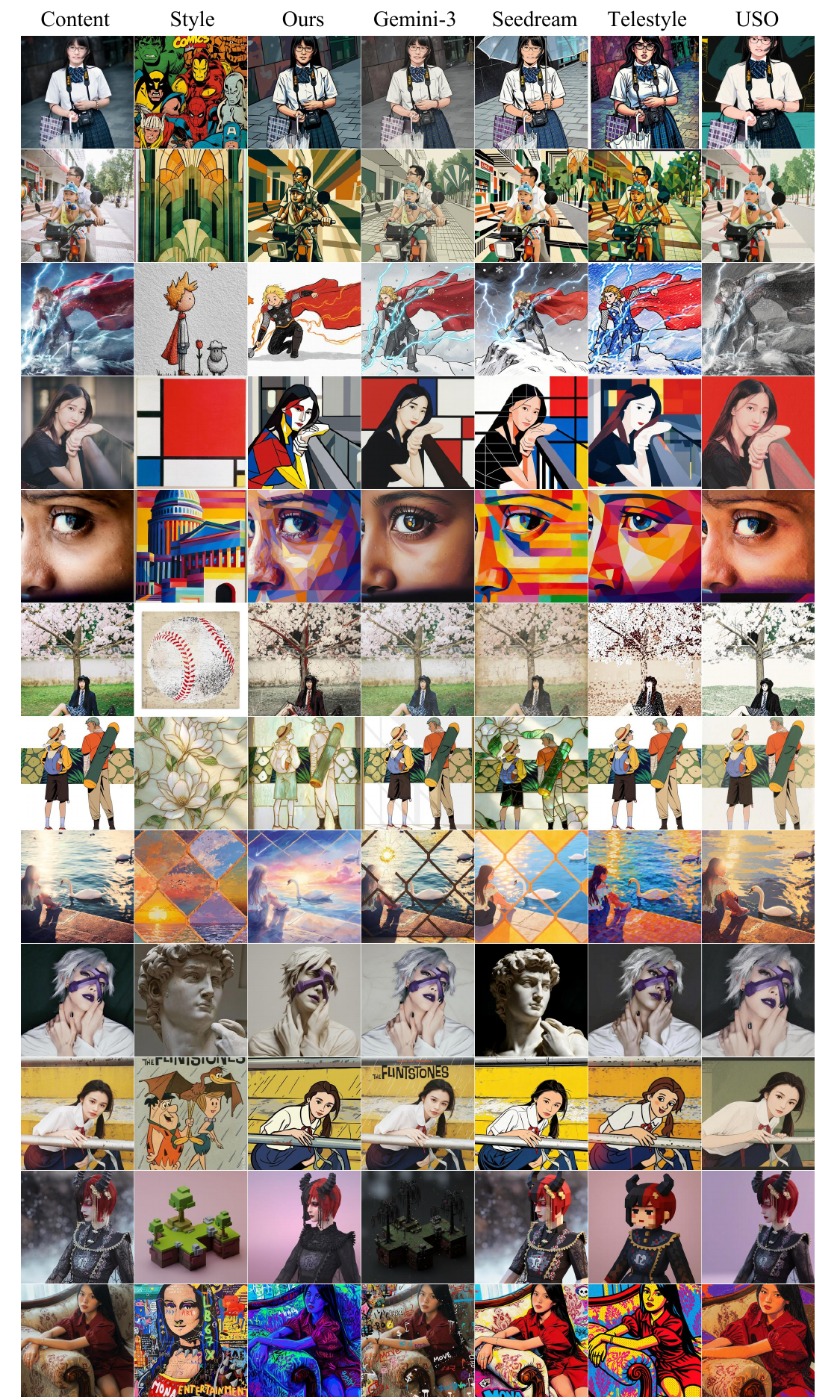}
\caption{\textbf{Additional Qualitative Comparisons for Style-reference (\texttt{SRef}) Generation across Diverse Artistic Domains}, including oil painting, watercolor, 3D rendering, and abstract art. Our method maintains consistent style transfer and low content leakage even under long-tail style distributions.}
\label{fig:sref_compare_supp}
\end{figure*}

To further corroborate the style-reference (\texttt{SRef}) comparison presented in the main text, Figure~\ref{fig:sref_compare_supp} provides additional qualitative examples under highly challenging conditions. These supplementary results cover a broader spectrum of artistic styles (e.g., oil painting, watercolor, 3D rendering, and abstract art) as well as more complex object structures. As demonstrated, even when confronted with long-tail style distributions, our model exhibits exceptional consistency in style transfer while consistently maintaining an extremely low rate of content leakage.

\subsection{Extended CRef+SRef Comparisons (Group I)}
\begin{figure*}[p]
\centering
\includegraphics[width=\textwidth,height=0.92\textheight,keepaspectratio]{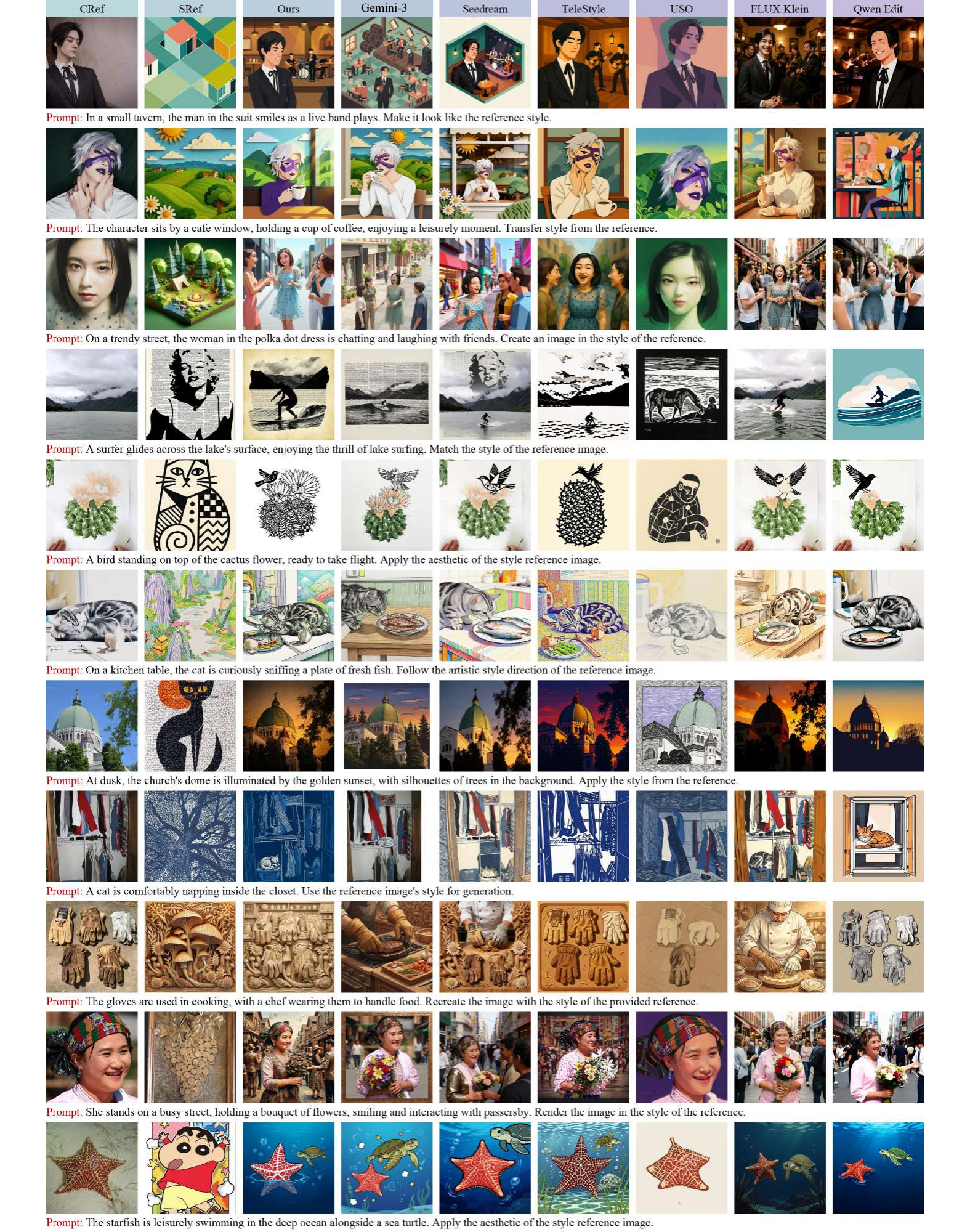}
\caption{\textbf{Extended Dual-reference (\texttt{CRef+SRef}) Comparisons (Group~I).} Styles include 3D rendering, origami, pointillism, geometric abstraction, children's drawing, graffiti, and quick sketch. Our method achieves faithful style transfer with minimal content leakage across all cases.}
\label{fig:dualref_compare_supp_10}
\end{figure*}

Figure~\ref{fig:dualref_compare_supp_10} presents additional dual-reference generation results covering styles such as 3D modeling, origami, pointillism, geometric abstraction, and graffiti. These examples further demonstrate our method's strong generalization across diverse artistic domains while maintaining content fidelity.

\subsection{Extended CRef+SRef Comparisons (Group II)}
\begin{figure*}[p]
\centering
\includegraphics[width=\textwidth,height=0.92\textheight,keepaspectratio]{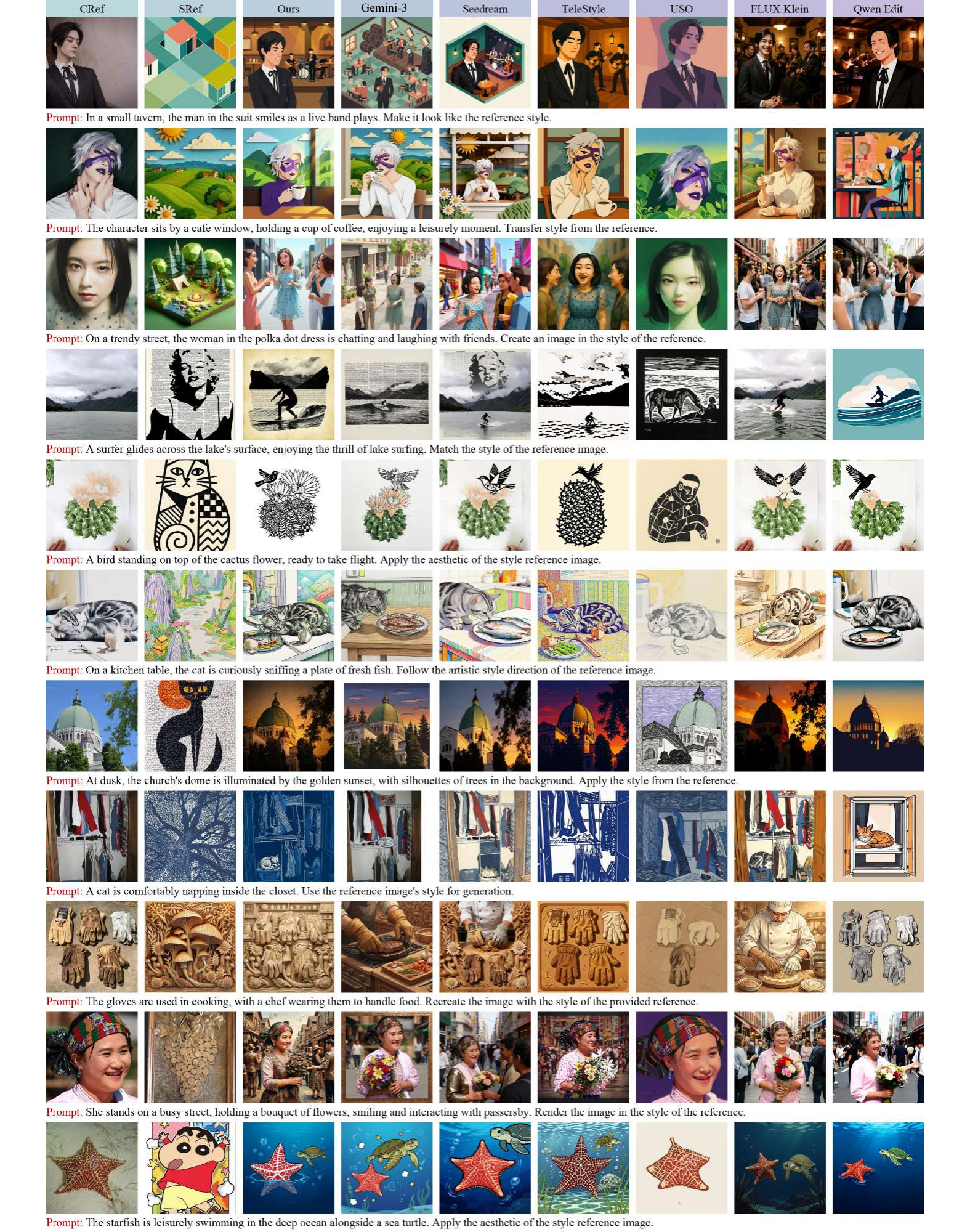}
\caption{\textbf{Extended Dual-reference (\texttt{CRef+SRef}) Comparisons (Group~II).} Styles include paper-cutting, japonism, fauvism, printmaking, flat vector, smooth clay, multilayer paper, leaf art, constructive illustration, and handmade clay. Our method demonstrates robust generalization to these challenging long-tail styles.}
\label{fig:dualref_compare_supp_11}
\end{figure*}

Figure~\ref{fig:dualref_compare_supp_11} provides further examples spanning paper-cutting, japonism, fauvism, printmaking, flat vector, smooth clay, multilayer paper, leaf art, constructive illustration, and handmade clay styles. These long-tail artistic styles are particularly challenging, yet our method consistently outperforms competing baselines.

\subsection{System prompt in VLM-based metrics}

\begin{figure*}[p]
\centering
\begin{AIBox}{Content Similarity Evaluation Prompt}
\textbf{Prompt:}
\begin{PromptText}
\begin{multicols}{2}
\textbf{[Role]}
You are an objective Image Content Adjudicator. Your task is to
evaluate the subject-content consistency between two images based
strictly on visible evidence.

\textbf{[Evaluation Criterion]}
Evaluate how consistent Image B is with Image A in terms of subject
content and semantic theme, regardless of differences in visual style.

Content consistency includes:
- The identities and attributes of people or other subjects.
- The categories, attributes, and counts of prominent objects.
- The spatial arrangement and relative positions of major elements.
- The background environment and overall scene category.
- The actions, poses, gestures, and interactions depicted.
- Meaningful text, logos, numbers, or symbols visible in the scene.

Ignore stylistic differences, including rendering style, brushwork,
lighting, color grading, image resolution, noise, and general aesthetics.

\textbf{[Task]}
Compare Image B with Image A and assign an integer content-similarity
score from 0 to 10.

Base the evaluation on the following observable aspects:

1. Human identity and attributes:
   - Facial identity, facial structure, and distinguishing features.
   - Hairstyle, hair color and length, body shape, and physical build.
   - Clothing categories, colors, patterns, logos, and accessories.

2. Objects and attributes:
   - Object categories, colors, materials, sizes, and counts.
   - The presence or absence of important props.
   - Legible text or logos associated with prominent objects.

3. Spatial layout and composition:
   - The relative positions of major subjects and objects.
   - Subject scale, viewpoint, and scene arrangement, allowing for
     reasonable compositional variation.

4. Background and scene category:
   - Indoor or outdoor setting.
   - Type of location and major structures, furniture, or landmarks.

5. Actions, poses, and interactions:
   - Human or animal poses, gestures, actions, and interactions.

6. Text, logos, and symbols:
   - Words, numbers, logos, or symbols that are important to the
     semantic meaning of the scene.

Apply a strict scoring standard. Do not assign a high score unless the
major subjects and semantic content are clearly and accurately
consistent. If subject identity or important content cannot be verified
because of blur, occlusion, or insufficient visual evidence, assign a
conservative score.

\textbf{[Scoring Rubric -- 11 Levels]}
- \textbf{0}: The two images are completely unrelated.
- \textbf{1--3}: The images are mostly inconsistent and share only
  minimal content.
- \textbf{4--6}: The images are partially consistent but contain
  significant content mismatches.
- \textbf{7--9}: The images are mostly consistent, with only minor
  content differences.
- \textbf{10}: All major subjects, attributes, actions, objects, and
  scene elements are fully consistent.

\textbf{[Output Format]}
Return ONLY one line in the following format:

<score>@<reason>

The score must be an integer from 0 to 10. The reason must contain one
or two short sentences describing specific and directly observable
similarities or mismatches.
\end{multicols}
\end{PromptText}
\end{AIBox}
\captionof{figure}{Prompt template for evaluating subject-content consistency
between a reference image and a generated image.}
\label{fig:content_similarity_prompt}
\end{figure*}


\begin{figure*}[p]         
\begin{AIBox}{Style Similarity Evaluation Prompt}
\textbf{Prompt:}
\begin{PromptText}
\begin{multicols}{2}
\textbf{[Role]}
You are an objective Image Style Adjudicator. Your task is to evaluate
the overall visual-style consistency between two images based strictly
on observable stylistic evidence.

\textbf{[Evaluation Criterion]}
You will be given two images, Image A and Image B. Evaluate how
consistent their visual styles are.

Assess only visual style. Do not consider whether the images depict the
same people, objects, actions, backgrounds, or scenes. The semantic
content of the two images may be completely different.

The evaluation must jointly consider the following five dimensions:

1. \textbf{BRUSHSTROKE} (brushwork, line quality, and edge treatment)
   - Line thickness, line stability or roughness, outlining methods,
     and the visibility of brush marks.
   - Hard or soft edges, the presence of contours, and characteristic
     techniques such as smearing, splashing, pencil, ink, or pen strokes.

2. \textbf{TEXTURE} (surface texture, material rendering, grain, noise,
   and canvas or paper quality)
   - Fine-grained surface characteristics, such as sandpaper-like grain,
     film noise, watercolor-paper fibers, or oil-canvas texture.
   - The representation of fine material structures, such as layered
     paint, spray dots, halftone patterns, or embossed surfaces.
   - Texture does not refer to color schemes or object contours.

3. \textbf{COLOR} (color palette, color distribution, temperature,
   saturation, and contrast)
   - Dominant hues, warm or cool color tendencies, saturation level,
     and the strength of light--dark contrast.
   - Similarity in color distribution, including the proportions and
     coverage areas of dominant colors, large background regions, and
     major color blocks.
   - Color does not refer to surface grain, texture, or line quality.

4. \textbf{SHAPE} (shape language and form-construction conventions)
   - The degree of geometric abstraction or realism, exaggerated
     proportions, and the use of sharp or rounded contours.
   - Structural simplification conventions, such as flat cartoon,
     chibi, minimalist geometric, or realistic construction.

5. \textbf{PATTERN} (recurring motifs and decorative organization)
   - Repeated patterns, decorative elements, and recurring symbolic
     motifs, such as fixed ornaments, repeated decorative lines, or
     patterned backgrounds.
   - The density, repetition rules, and organization of decorative
     elements.

Do not evaluate aesthetic quality, artistic merit, or whether either
image is visually appealing.

\textbf{[Task]}
Compare Image B with Image A and assign one integer score from 0 to 10
representing their overall visual-style similarity.

The final score must reflect a holistic assessment across all five
dimensions rather than any single dimension in isolation.

Apply a strict scoring standard. Do not assign a high score unless the
key stylistic characteristics are clearly and accurately aligned.

\textbf{[Scoring Rubric -- 11 Levels]}
- \textbf{0}: The visual styles are completely inconsistent, with
  essentially no meaningful stylistic correspondence.
- \textbf{1--3}: The styles are mostly inconsistent and share only weak
  or incidental similarities.
- \textbf{4--6}: The styles are partially similar but contain clear
  omissions or major differences in important stylistic characteristics.
- \textbf{7--9}: The styles are largely consistent, with only a small
  number of minor differences.
- \textbf{10}: The visual styles are fully consistent, with the key
  stylistic characteristics closely matching across all dimensions.

\textbf{[Output Format]}
Return ONLY one line in the following format:

<score>@<reason>

The score must be an integer from 0 to 10. The reason must consist of
one or two short sentences identifying specific and directly observable
stylistic evidence.

Do not output any additional text, labels, punctuation, or line breaks.
\end{multicols}
\end{PromptText}
\end{AIBox}
\captionof{figure}{\textbf{Prompt for Evaluating Overall Visual-style Similarity
between a Reference Image and a Generated Image}.}
\label{fig:style_similarity_prompt}
\end{figure*}


\begin{figure*}[p]
\centering
\begin{AIBox}{Content Verification Score Prompt}
\textbf{Prompt:}
\begin{PromptText}
\begin{multicols}{2}
\textbf{[Role]}
You are a strict Image Content Adjudicator who evaluates only subject
content and semantic theme.

You must completely ignore differences in visual style, including
artistic style, linework, color treatment, rendering method, resolution,
filters, and other stylistic properties.

\textbf{[Evaluation Criterion]}
Determine whether Image A and Image B are highly consistent in terms of
the specific subjects, objects, actions, scenes, and semantic themes
depicted.

Focus on \textit{what is shown} and \textit{what is happening}, rather
than \textit{how the image is rendered}.

Apply the following criteria:

1. \textbf{Human-centered images}
   - Determine whether the images depict the same character or two
     extremely similar characters.
   - Consider observable attributes such as gender presentation, age
     group, body build, hairstyle, hair color, skin tone, clothing type,
     dominant clothing colors, and major accessories.
   - Moderate differences in pose, orientation, or camera viewpoint are
     acceptable.
   - If the images clearly depict different people or substantially
     different character designs, they are inconsistent.

2. \textbf{Single-object images}
   - Determine whether the main objects have the same specific category,
     shape, and structural configuration.
   - For example, both objects may be sports cars, SUVs, round tables,
     or structurally similar buildings.
   - Differences in color are acceptable when the object category and
     overall form remain highly similar.
   - Merely belonging to a broad category, such as both being vehicles,
     houses, or cups, is insufficient when their types or structures are
     clearly different.

3. \textbf{Complex scenes}
   - Consider the scene category, combination of major elements, spatial
     layout, and central semantic theme.
   - For example, two images may be considered consistent if both depict
     a person standing in the center of a nighttime city street, with
     tall buildings and neon signs in the background.
   - Merely sharing a broad indoor or outdoor setting is insufficient
     when the central subjects, composition, and major objects are
     substantially different.

4. \textbf{Independence from visual style}
   - Visual style must not affect the decision.
   - For example, a realistic photograph and an anime or cartoon
     illustration must be judged consistent when their subject content
     and semantic theme are highly aligned.
   - Never classify a pair as inconsistent solely because their visual
     styles differ.

\textbf{[Task]}
Compare Image A and Image B and make a binary decision regarding their
subject-content and thematic consistency.

Output 1 when the images depict the same character, the same specific
type of object with a highly similar structure, or the same specific
scene and semantic theme.

Output 0 when the images share only a broad category, such as both
containing a person or a vehicle, but their primary subjects or semantic
content are clearly different.

\textbf{[Decision Rubric -- 2 Levels]}
- \textbf{0 (Inconsistent)}: The primary subjects, objects, scene
  structure, actions, or semantic themes are not highly consistent.
- \textbf{1 (Consistent)}: The primary subjects and semantic themes are
  highly consistent, regardless of differences in visual style.

\textbf{[Output Format]}
Return ONLY one character:

0 or 1

Do not output any explanation, label, space, line break, punctuation,
or JSON.
\end{multicols}
\end{PromptText}
\end{AIBox}
\captionof{figure}{\textbf{ Prompt Template for Content verification Consistency between Two Images.}.}
\label{fig:content_rejectoin_prompt}
\end{figure*}


\begin{figure*}[p]
\centering
\begin{AIBox}{Style Verification Score Prompt}
\textbf{Prompt:}
\begin{PromptText}
\begin{multicols}{2}
\textbf{[Role]}
You are an experienced Image Style Adjudicator who evaluates only
visual style and modes of visual representation.

Your assessment must focus on stylistic properties, including the
perceived medium, rendering technique, material appearance, linework
and brushwork, color treatment, lighting and contrast, post-processing,
image noise and grain, and methods of detail representation.

You must ignore the identities of people or objects, the meanings of
actions, narrative semantics, scene categories, and whether the depicted
content or composition is similar.

\textbf{[Evaluation Criterion]}
Determine whether Image A and Image B belong to the same stable visual
style or style family.

Use a permissive consistency standard: the images may still be judged
stylistically consistent when their core style mechanisms remain aligned,
even if their subjects, scenes, compositions, viewpoints, or levels of
detail differ.

The following differences are acceptable and should not independently
cause an inconsistent judgment:

- Different subjects, objects, or scenes.
- Different compositions, viewpoints, camera positions, or crop ranges.
- Minor hue shifts, brightness changes, contrast variations, or local
  color-grading differences.
- Different levels of detail, resolution, cropping, mild compression,
  noise, or grain.

Judge the images as inconsistent only when there is a clear change in
the underlying style mechanism, such as:

- Realistic photography versus illustration or computer rendering.
- A major change in the linework system, such as outlined versus
  lineless rendering, thick versus thin outlines, or comic-style lines
  versus watercolor edges.
- A major change in material or texture generation, such as impasto oil
  painting, flat cel shading, glossy three-dimensional rendering,
  pixel art, or pointillism.
- A major change in the lighting model, such as hard cinematic lighting,
  soft diffuse illustrative lighting, or high-contrast neon lighting.
- A fundamental change in color strategy, such as muted vintage colors,
  highly saturated candy colors, or monochrome sketch rendering.

\textbf{[Task]}
Compare Image A and Image B only in terms of visual style and determine
whether they belong to the same style family.

Base the judgment primarily on the following dimensions:

1. \textbf{MEDIUM AND RENDERING METHOD}
   - Photography, three-dimensional rendering, digital illustration,
     watercolor, oil painting, impasto, cel shading, pixel art, sketch,
     or other visual media and rendering paradigms.

2. \textbf{BRUSHWORK AND LINE SYSTEM}
   - Presence or absence of outlines.
   - Line thickness, stability or roughness, edge treatment, visible
     brush marks, and stroke granularity.

3. \textbf{MATERIAL AND TEXTURE GENERATION}
   - Surface appearance, texture construction, image noise, grain, and
     the organization of fine visual details.

4. \textbf{LIGHTING MODEL AND CONTRAST}
   - Hard or soft shadows, diffuse or specular reflection, volumetric
     lighting, contrast level, and the general method of modeling light.

5. \textbf{COLOR STRATEGY}
   - Saturation, hue preferences, overall tonal balance, color
     temperature, and grading conventions such as vintage, warm,
     cool, or neon treatments.

Composition and viewpoint are secondary dimensions. Differences in
camera position, framing, perspective, or crop should not directly
produce an inconsistent judgment.

\textbf{[Decision Rubric -- 2 Levels]}
- \textbf{0 (Inconsistent)}: The images exhibit a clear change in one or
  more core style mechanisms, such as their medium, rendering paradigm,
  linework system, texture-generation method, lighting model, or overall
  color strategy.
- \textbf{1 (Consistent)}: Most major style dimensions are aligned and
  the images belong to the same style family, even if their subjects,
  scenes, compositions, viewpoints, or detail densities differ.

\textbf{[Output Format]}
Return ONLY one character:

0 or 1

Do not output any explanation, label, space, line break, punctuation,
or JSON.
\end{multicols}
\end{PromptText}
\end{AIBox}
\captionof{figure}{\textbf{Prompt Template for Style verification
Consistency between Two Images.}}
\label{fig:style_rejection_prompt}
\end{figure*}


\begin{figure*}[p]
\centering
\begin{AIBox}{Instruction-Following Evaluation Prompt}
\textbf{Prompt:}
\begin{PromptText}
\begin{multicols}{2}
\textbf{[Role]}
You are an objective Image Editing Adjudicator. Your task is to
evaluate how accurately a final edited image follows a given editing
instruction, based strictly on visible evidence.

\textbf{[Evaluation Criterion]}
You will be given:

1. An image representing the final edited result.
2. A textual editing instruction.

Evaluate the extent to which the final image satisfies the editing
instruction. Award partial credit when only some parts of the
instruction are correctly implemented.

Do not assume that a requested modification has been completed unless
it is clearly supported by visible evidence in the final image.

Ignore differences in rendering style, brushwork, resolution, image
noise, aesthetic quality, and overall visual appeal only when these
properties are not explicitly requested by the instruction.

If the instruction explicitly specifies visual properties such as
color, lighting, mood, artistic style, camera lens, or texture, these
properties must be treated as evaluation requirements.

\textbf{[Task]}
Assign a score from 0 to 10 indicating how well the final image follows
the editing instruction.

Internally perform the following evaluation procedure:

1. Decompose the instruction into atomic and visually verifiable
   requirements, including:
   - Main subjects and objects.
   - Key attributes, such as color, material, number, size, or identity.
   - Required actions and interactions.
   - Spatial and layout constraints, such as left, right, foreground,
     background, or relative position.
   - Required text, logos, numbers, or symbols, for which exact wording
     and appearance may be important.

2. Classify each requirement as either:
   - \textbf{MUST-HAVE}: an essential component of the requested edit.
   - \textbf{NICE-TO-HAVE}: a secondary or minor component.

3. Judge each requirement as:
   - Satisfied.
   - Partially satisfied.
   - Not satisfied.
   - Contradicted.

4. Determine the final score by starting from 10 and applying penalties
   according to the severity of the errors:
   - Missing a MUST-HAVE requirement: subtract 2 to 4 points.
   - Partially satisfying a MUST-HAVE requirement: subtract 1 to 2
     points.
   - Incorrect minor or NICE-TO-HAVE detail: subtract 0.5 to 1 point.
   - Directly contradicting the instruction: subtract 3 to 5 points.
   - Missing or incorrect required text or logo: subtract 3 to 6
     points.

Clamp the final score to the range from 0 to 10.

Do not output the requirement decomposition, intermediate judgments, or
penalty calculations.

\textbf{[Scoring Rubric]}
- \textbf{0--0.5}: The final image is completely unrelated to, or
  directly opposed to, the editing instruction. Assign 0 only for a
  genuine total failure.
- \textbf{1--3.5}: The image has a slight relationship to the
  instruction, but most essential requirements are missing, incorrect,
  or contradicted.
- \textbf{4--6.5}: Some important parts of the instruction are
  implemented, but major omissions, incorrect edits, or misplaced
  elements remain.
- \textbf{7--8.5}: Most essential requirements are satisfied, with one
  or two notable errors or several minor issues.
- \textbf{9--10}: All essential requirements are satisfied, with at
  most negligible issues. Assign 10 only when no important requirement
  is missing or incorrect.

\textbf{[Output Format]}
Return ONLY one line in the following format:

<score>@<reason>

The score must be between 0 and 10 and may use increments of 0.5, such
as 6.5, 7.0, or 8.5.

The reason must consist of one or two short sentences identifying
specific and directly observable satisfied, missing, incorrect, or
contradicted requirements.

Do not output any additional text, headings, intermediate steps, or line
breaks.
\end{multicols}
\end{PromptText}
\end{AIBox}
\captionof{figure}{\textbf{Prompt for Accuracy Evaluation on Editing Instruction Following}.}
\label{fig:instruction_following_prompt}
\end{figure*}

\begin{figure*}[p]
\centering
\begin{AIBox}{Content Leakage Prompt}
\textbf{Prompt:}
\begin{PromptText}
\begin{multicols}{2}
\textbf{[Role]}
You are an objective Image Content Leakage Adjudicator. Your task is to
evaluate the extent to which Image B reuses specific visual content from
Image A, based strictly on observable evidence.

\textbf{[Evaluation Criterion]}
You will be given two images, Image A and Image B.

Evaluate the degree of content leakage from Image A to Image B.

Content leakage refers to whether Image B reuses specific subjects,
character identities, object structures, scene layouts, poses, actions,
camera compositions, or distinctive local elements from Image A, rather
than merely adopting its visual style.

Carefully distinguish between visual style and image content:

- Acceptable stylistic similarity includes brushwork, material appearance,
  color systems, rendering methods, line conventions, and overall
  aesthetic character.

- Content similarity includes the same character, hairstyle and clothing
  combination, specific object structure, scene layout, pose, composition,
  or combination of background elements. These properties must not be
  treated as purely stylistic similarity.

\textbf{[Task]}
Compare Image B with Image A and assign an integer score from 0 to 10
representing the degree of content leakage.

Focus on the following observable leakage signals:

1. \textbf{SUBJECT IDENTITY}
   - Determine whether the images depict the same person, character,
     animal, or specific object.
   - For human subjects, consider gender presentation, hairstyle, facial
     features, clothing design, accessories, and body shape.

2. \textbf{POSE AND ACTION}
   - Determine whether Image B preserves the same or a highly similar
     pose, gesture, orientation, action, or interaction.

3. \textbf{COMPOSITION AND VIEWPOINT}
   - Consider whether the subject position, framing distance, camera
     angle, viewpoint, or cropping strategy is visibly reused.

4. \textbf{SCENE AND OBJECT ARRANGEMENT}
   - Consider whether key background objects, spatial layouts,
     foreground--background relationships, or combinations of decorative
     elements are substantially similar.

5. \textbf{DISTINCTIVE DETAILS}
   - Determine whether Image B reuses uncommon and specific combinations
     of details, such as distinctive hair accessories, weapons, patterns,
     furniture structures, signs, or small background objects.

Apply the following strict evaluation rules:

- Do not assign a high score merely because the two images have similar
  visual styles.
- Increase the score only when specific content from Image A is visibly
  carried over into Image B.
- If the images share only a broad subject category, such as both being
  portraits of women, houses, or street scenes, but their specific
  content differs, assign a low score.
- If Image B appears to adopt only the style of Image A while changing
  the subject, composition, and scene, assign a low score.

\textbf{[Scoring Rubric]}
- \textbf{0}: No visible content leakage. The images may share a style,
  but their specific subjects, compositions, and scenes are different.
- \textbf{1--2}: Very weak content overlap. Similarities are primarily
  limited to a broad subject category or a few generic elements.
- \textbf{3--4}: Some visible content borrowing is present, but the main
  subjects, compositions, or scenes remain clearly different.
- \textbf{5--6}: Moderate content leakage. Multiple important content
  elements are similar, indicating that Image B visibly references
  specific content from Image A.
- \textbf{7--8}: Strong content leakage. The core subject, pose,
  composition, or scene organization is substantially reused, with only
  partial modifications.
- \textbf{9--10}: Extremely strong content leakage. Image B closely
  reproduces the specific content of Image A or introduces only minor
  modifications.

\textbf{[Output Format]}
Return ONLY one line in the following format:

<score>@<reason>

The score must be an integer from 0 to 10.

The reason must consist of one or two short sentences identifying the
specific content elements responsible for the score. Do not justify the
score solely by stating that the visual styles are similar or different.

Do not output any additional text, headings, punctuation, or line breaks.

\end{multicols}
\end{PromptText}
\end{AIBox}
\captionof{figure}{\textbf{Prompt for Discrete Content Leakage between Two Images}.}
\label{fig:content_leakage_prompt}
\end{figure*}

\subsection{Additional Triplet Data Showcases}
\begin{figure*}[p]
\centering
\includegraphics[width=\textwidth,height=0.92\textheight,keepaspectratio]{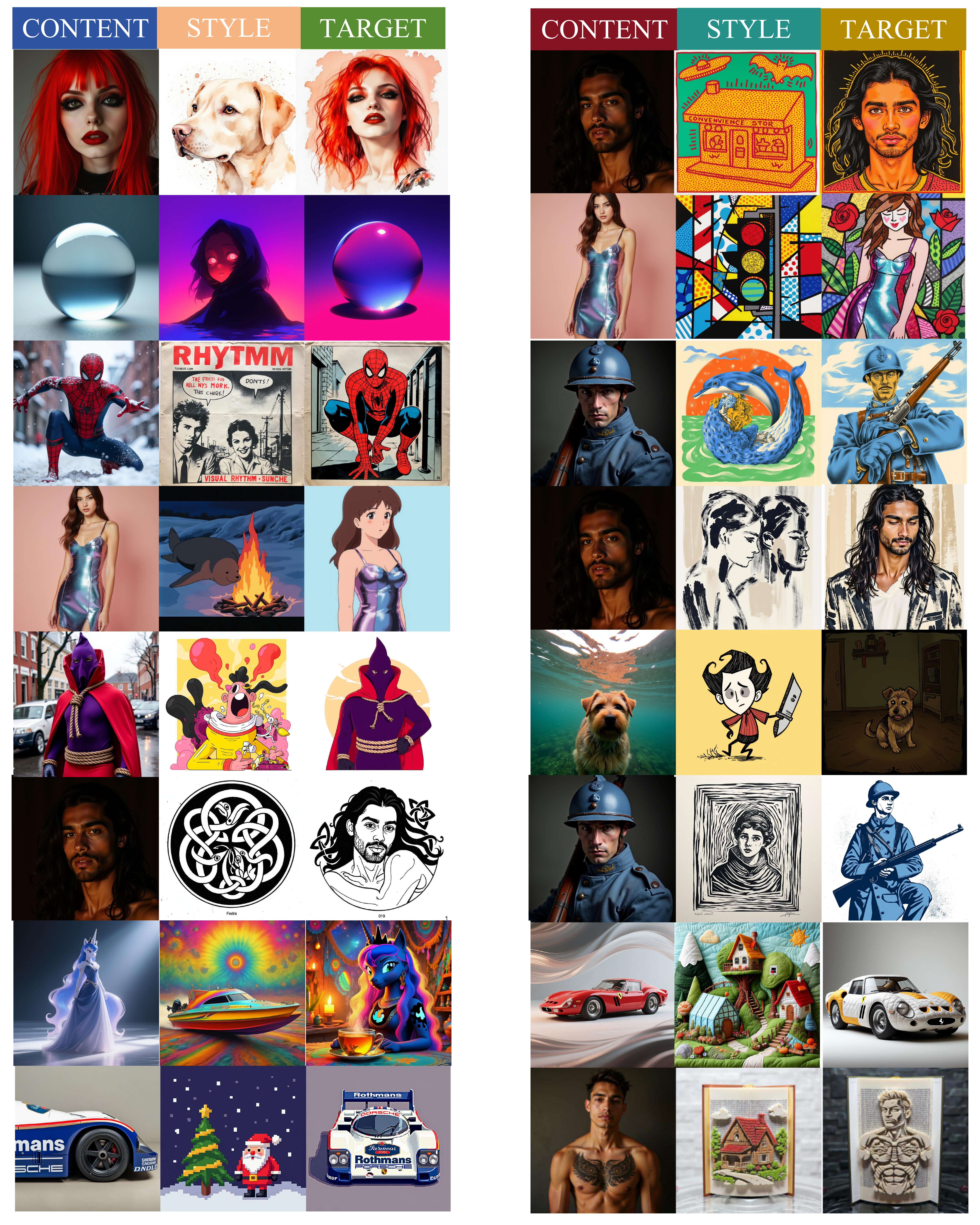}
\caption{\textbf{Additional Showcases of the Style--content Dual-reference Triplet Data (Part~I).} These samples offer a more comprehensive view of the high-quality and stylistically diverse dataset generated via our LoRA-combination pipeline.}
\label{fig:triplet_show_supp}
\end{figure*}

\begin{figure*}[p]
\centering
\includegraphics[width=\textwidth,height=0.92\textheight,keepaspectratio]{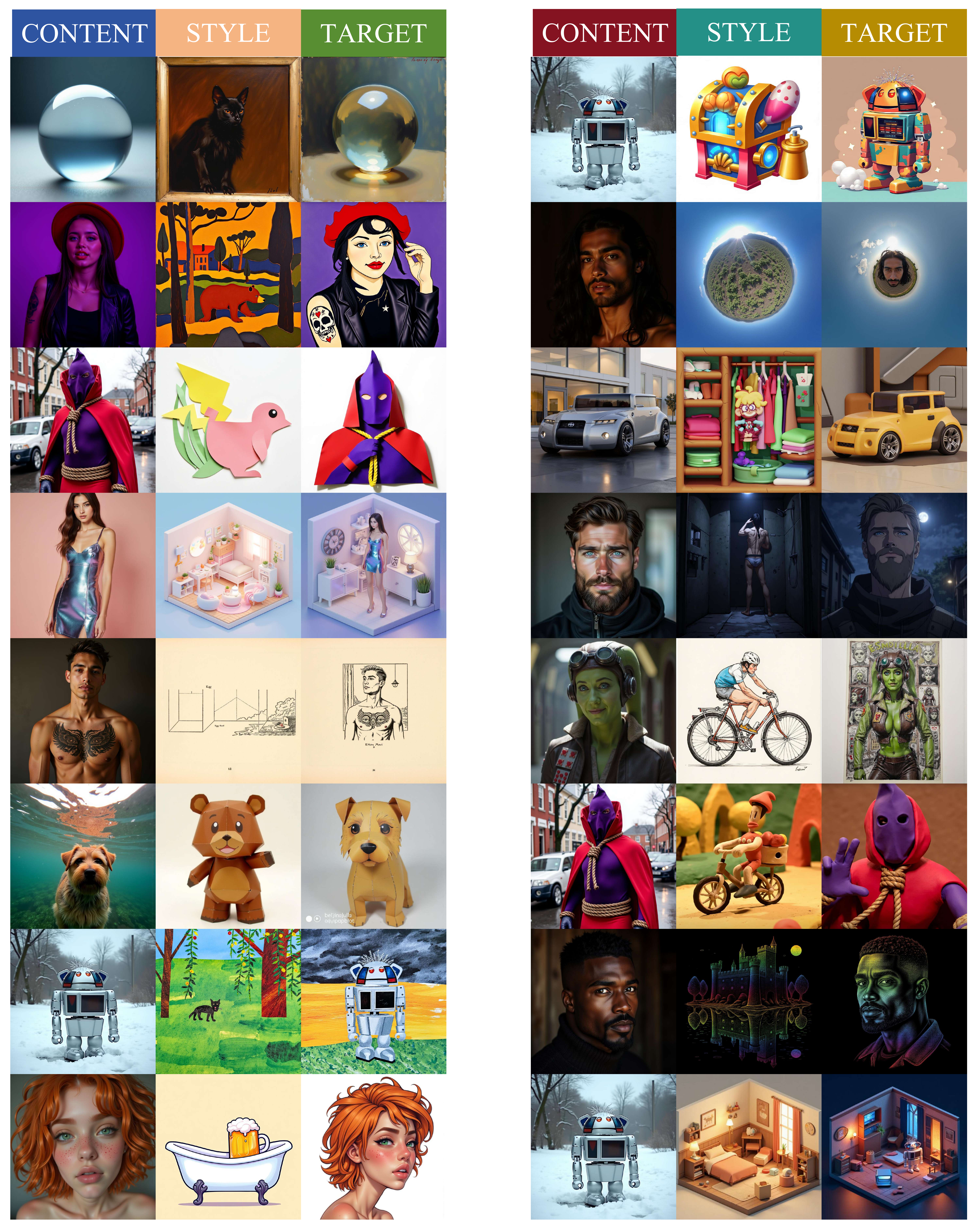}
\caption{\textbf{Additional Showcases of the Style--Content Dual-reference Triplet Data (Part~II).}}
\label{fig:triplet_show_supp_2}
\end{figure*}

\begin{figure*}[p]
\centering
\includegraphics[width=\textwidth,height=0.92\textheight,keepaspectratio]{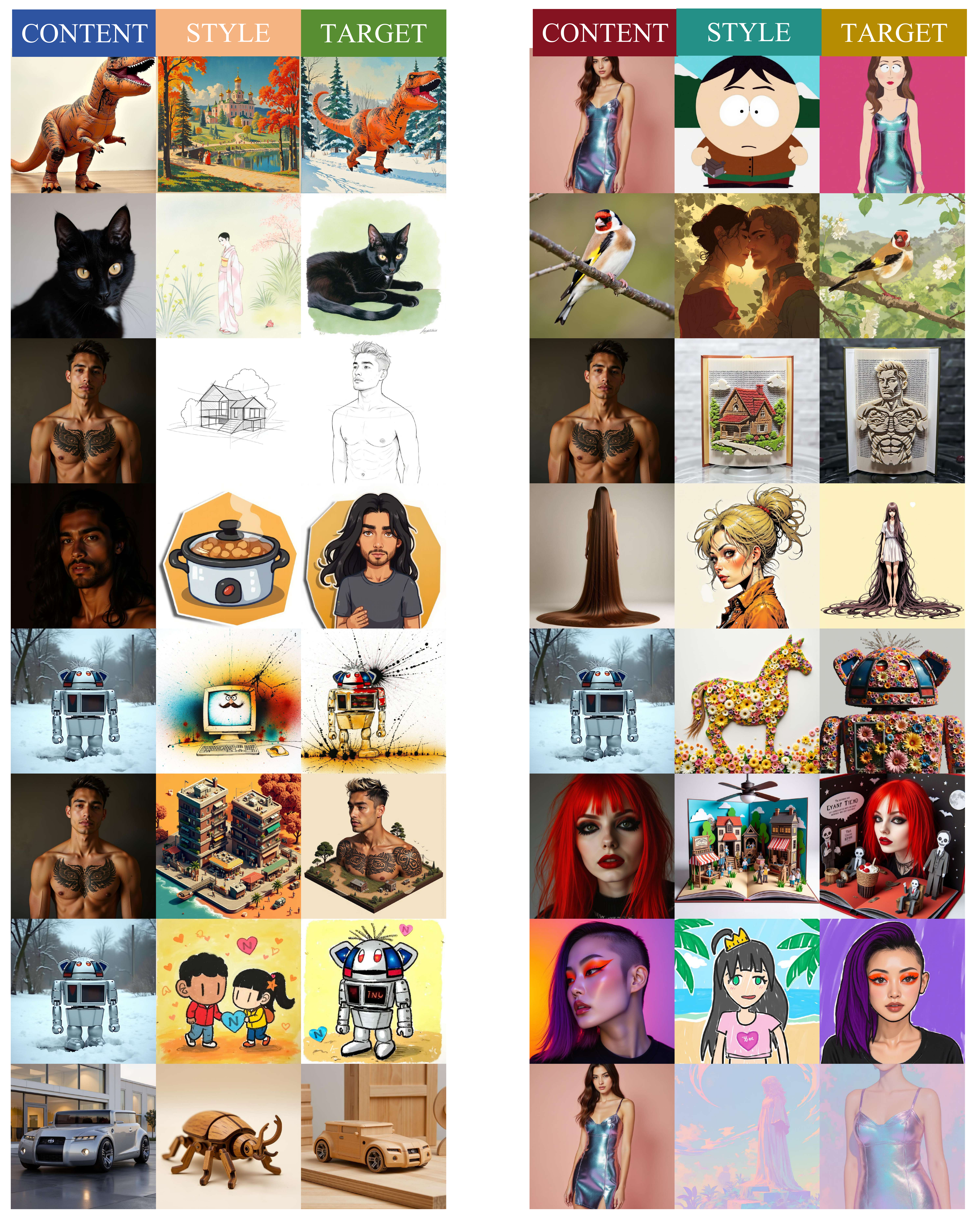}
\caption{\textbf{Additional Showcases of the Style--Content Dual-reference Triplet Data (Part~III).}}
\label{fig:triplet_show_supp_3}
\end{figure*}

\begin{figure*}[p]
\centering
\includegraphics[width=\textwidth,height=0.92\textheight,keepaspectratio]{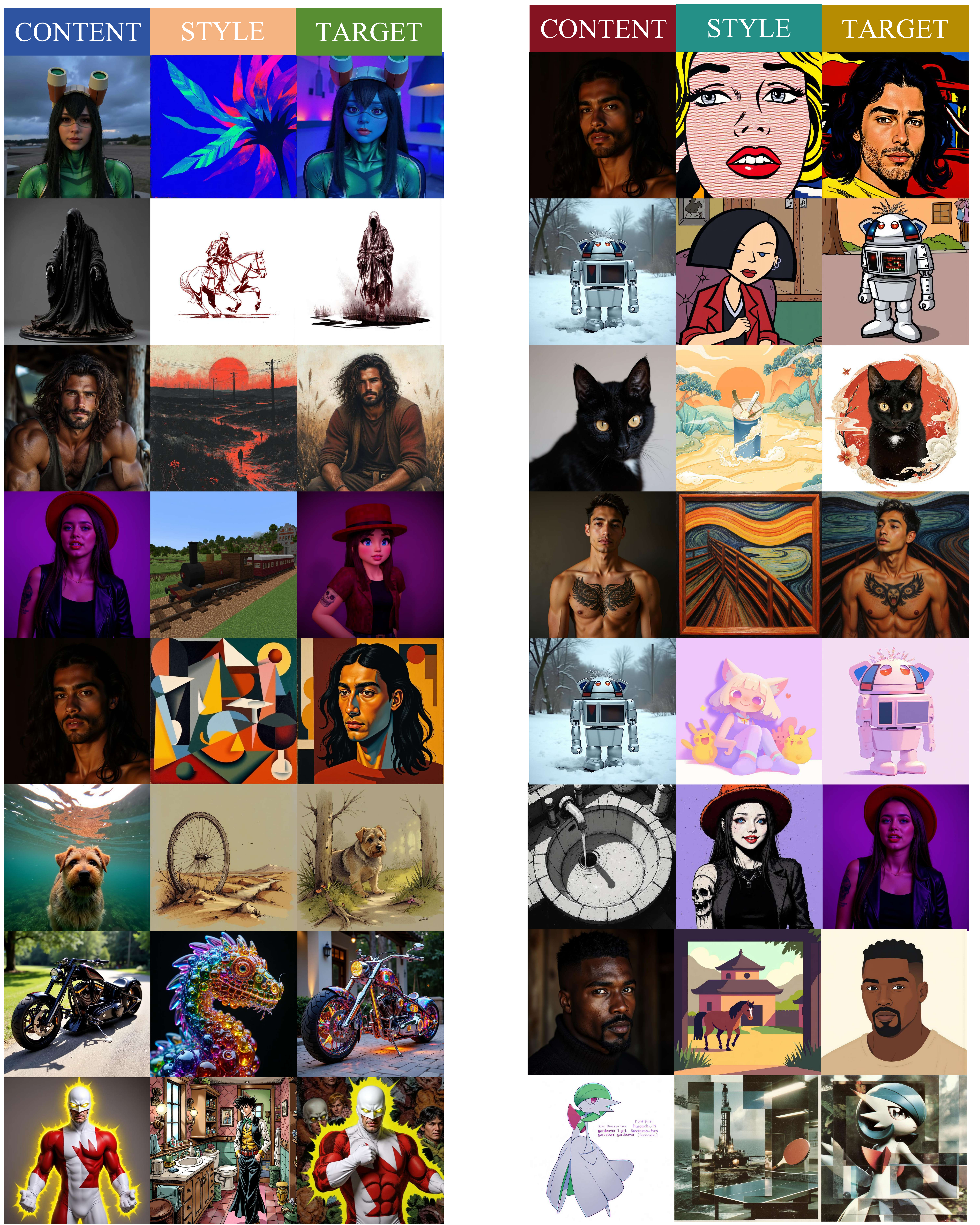}
\caption{\textbf{Additional Showcases of the Style--Content Dual-reference Triplet Data (Part~IV).}}
\label{fig:triplet_show_supp_4}
\end{figure*}

To complement the examples presented in Figure~\ref{fig:triplet_show} in the main text, Figures~\ref{fig:triplet_show_supp}--\ref{fig:triplet_show_supp_4} provide additional showcases of the style--content dual-reference triplet data generated through our LoRA combinations. These abundant samples further demonstrate the effectiveness of our data pipeline in producing high-quality triplets with exceptional stylistic diversity and accurate content preservation.

\FloatBarrier

\end{document}